\documentclass[preprint,10pt]{elsarticle}
\usepackage[a4paper,top=2cm,bottom=2cm,left=2.5cm,right=2.5cm,marginparwidth=1.75cm]{geometry}
\usepackage{amsmath}
\usepackage{graphicx} 
\usepackage{makecell}
\usepackage{multirow}
\usepackage[table]{xcolor}
\usepackage[title]{appendix}
\usepackage{float}
\usepackage[section]{placeins}
\usepackage{amsmath}
\usepackage{bm}
\usepackage{paralist}
\usepackage{microtype}
\usepackage{algorithm}
\usepackage{algpseudocode}
\usepackage{url}
\usepackage{natbib}
\usepackage{orcidlink}
\usepackage[nocompress]{cite}

\usepackage[labelfont=bf]{caption}
\journal{arXiv preprint}

\begin{document}

\begin{frontmatter}

\title{Deep Learning-Based Multi-Object Tracking: A Comprehensive Survey from Foundations to State-of-the-Art}

\author[1]{Momir Adžemović~\href{https://orcid.org/0009-0001-1385-4094}{\includegraphics[scale=0.055]{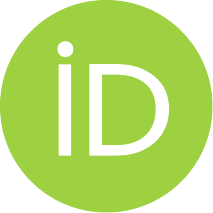}}}\ead{pd222011@alas.matf.bg.ac.rs}\cortext[cor1]{Corresponding author}

\affiliation[1]{organization={Department of Computer Science, Faculty of Mathematics, University of Belgrade},
addressline={Studentski trg 16}, 
city={Belgrade},
postcode={105104}, 
country={Serbia}}

\begin{abstract}

Multi-object tracking (MOT) is a core task in computer vision that involves detecting objects in video frames and associating them across time. The rise of deep learning has significantly advanced MOT, particularly within the tracking-by-detection paradigm, which remains the dominant approach. Advancements in modern deep learning-based methods accelerated in 2022 with the introduction of ByteTrack for tracking-by-detection and MOTR for end-to-end tracking. Our survey provides an in-depth analysis of deep learning-based MOT methods, systematically categorizing tracking-by-detection approaches into five groups: joint detection and embedding, heuristic-based, motion-based, affinity learning, and offline methods. In addition, we examine end-to-end tracking methods and compare them with existing alternative approaches. We evaluate the performance of recent trackers across multiple benchmarks and specifically assess their generality by comparing results across different domains. Our findings indicate that heuristic-based methods achieve state-of-the-art results on densely populated datasets with linear object motion, while deep learning-based association methods, in both tracking-by-detection and end-to-end approaches, excel in scenarios with complex motion patterns.

\end{abstract}

\date{February 2025}


\end{frontmatter}

\section{Introduction}
\label{sec:introduction}

Multi-object tracking (MOT) is a computer vision task that involves detecting and tracking objects of a given category in a video. It operates without prior knowledge of the objects' appearance or their location. Compared to object detection applied to a video, it additionally requires associating detections across frames by assigning them unique object IDs. Solving MOT results in a set of tracks, where each track consists of all detections belonging to the same object. This task is challenging, especially when aiming for a general solution. The main difficulty arises from frequent occlusions, where MOT algorithms, i.e. trackers, may switch object IDs or lose track of some objects. Additionally, objects with similar appearances make it difficult to distinguish between them, while unpredictable dynamic motion further complicates tracking. Despite these challenges, MOT is essential in various fields that require real-time video analysis. Applications include autonomous driving~\citep{nuscenes, kitti, survey2021_autonomous_vehicles}, sports analysis~\citep{sportsmot, soccernet}, retail~\citep{retail_analytics}, robotics~\citep{object_tracking_in_robotics, robotics_top_view_surveillance}, biology~\citep{biology}, and surveillance~\citep{surveillance_systems_review, surveillance_systems_research}, among others.
The emergence of deep learning-based methods led to introduction of tracking-by-detection frameworks. This paradigm started with a pioneering work, Simple Online and Real-time Tracker (SORT)~\citep{sort}, demonstrated the effectiveness of using CNN object detectors along with Kalman filtering (KF) for motion prediction and Hungarian matching~\citep{linear_assignment} for detection association across frames. Deep SORT further improved tracking accuracy with the integration of deep learning-based appearance extraction in the association step for more robust tracking in complex scenarios. From that point onward, deep learning is stample in MOT. 

Deep learning-based methods in tracking-by-detection paradigm compared to the SORT pioneer algorithm have improved by multiple folds. First, object detection methods have improved by a large margin. New and improved versions of existing architecture families are released~\citep{yolov5yolov8yolov10goto, yolov10, yolov8review}. Additionally, novel transformer-based architectures are also now available~\citep{detr, rt_detr}. Second, association methods have improved in various aspects that include: improved appearance feature extractors~\citep{botsort, deepocsort}, integration of camera motion compensation (CMC) methods for camera movement robustness, improved motion models~\citep{motiontrack, ettrack, movesort}, and many association heuristics~\citep{bytetrack, ocsort, hybridsort, boostrack, sparsetrack, cbiou, deep_eiou}. Lastly, offline methods based on graph neural networks (GNN)~\citep{gnn_survey} have been introduced to solve the association with higher accuracy compared to online variants. 

More paradigms have emerged, capable of performing both detection and association in an end-to-end manner. One of such paradigms are tracking-by-query methods with MOTR~\citep{motr} and TrackFormer~\citep{trackformer} as the pioneers. These methods are based on the DETR~\citep{detr} end-to-end detection model. Compared to the tracking-by-detection methods, which often require domain knowledge and heuristics for strong performance, end-to-end methods are general solutions that are able to fully learn detection and association directly from data. 

With the rapid advancements in deep learning-based MOT, a systematic review is necessary to categorize existing approaches, highlight emerging trends, and identify open research challenges. This paper presents a comprehensive survey of modern deep learning-based algorithms. We specifically examine advancements in both tracking-by-detection and end-to-end paradigms, with an emphasis on state-of-the-art methods. To evaluate their performance across different domains, we compare these methods on multiple benchmarks containing 2D data collected from a single camera. Furthermore, we define a custom benchmark based on existing datasets to evaluate how well each tracking algorithm generalizes. Our contributions are as follows:

\begin{itemize}
    \item We present a detailed survey of modern deep learning-based methods, with an emphasis on recent works (i.e., from 2022) that have not been covered in detail in existing surveys. The survey is structured into two parts: the first part focuses on the tracking-by-detection paradigm, while the second covers end-to-end tracking methods.  
    \item We identify key categories within the tracking-by-detection paradigm and systematically categorize each reviewed method.  
    \item We compile experimental results from multiple benchmarks to analyze tracker performance across different domains. In addition, we evaluate tracker generality by aggregating and comparing results across benchmarks.  
    \item We provide a comprehensive background, ensuring accessibility for readers with diverse expertise.  
\end{itemize}

\section{Related work}

Several comprehensive surveys on multi-object tracking have been published in recent years~\citep{survey2019_deep_learing_in_MOT, survey2020_associaton_model_based, survey2021_deep_learning_review_by_components, survey2021_autonomous_vehicles, survey2022_traffic_environmnets, survey2024_embedding_methods, survey2022_association_methods, mot_gnn_survey}. We present an overview of these surveys along with their topics in Table~\ref{tab:related_review_papers}. We group these papers into \textit{general} and \textit{domain} categories. The general category consists of papers that review trackers without focusing on any specific domain or architecture. The domain category provides an overview of trackers applied to a specific domain.

\textbf{General literature surveys}. In~\citep{survey2019_deep_learing_in_MOT}, the authors review the application of deep learning to different tracker components: object detection, feature extraction, motion prediction, affinity computation, and data association. This survey primarily covers works from 2016 to 2019, with a strong focus on tracking-by-detection approaches, which dominated the field at the time. In~\citep{survey2021_deep_learning_review_by_components}, the authors examine the transition from traditional to deep learning-based methods, structuring their analysis around fundamental tracker components. In~\citep{survey2020_associaton_model_based}, the authors review machine learning-based data association methods, focusing on their connections to linear assignment and analyzing various optimization approaches. In~\citep{survey2022_association_methods}, they provide a broad overview of data association techniques, identifying effective methods while discussing challenges and open problems that future research should address. In~\citep{survey2024_embedding_methods}, the authors conduct a detailed review of embedding-based MOT approaches, categorizing different embedding techniques and evaluating their effectiveness for feature representation and association. This survey also summarizes widely used MOT datasets and examines the advantages of state-of-the-art embedding strategies. In~\citep{mot_gnn_survey}, the authors review graph-based data association methods in MOT, highlighting the role of graph structures in modeling object dependencies and analyzing recent deep learning implementations alongside traditional approaches.

\textbf{Domain literature surveys}. In~\citep{survey2021_autonomous_vehicles}, the authors review MOT approaches specifically designed for autonomous vehicles, analyzing challenges such as occlusions, varying lighting conditions, and real-time processing constraints. In~\citep{survey2022_traffic_environmnets}, they conduct a systematic literature review of MOT in traffic environments, covering techniques, datasets, and evaluation metrics while highlighting key challenges and future research directions.

\begin{table*}[h]
\small
\centering
\begin{tabular}{c l c c}
Reference & Topic & Group & Year \\
\hline
\citep{survey2019_deep_learing_in_MOT} & \makecell[l]{Deep learning in MOT: Overview of object detection models, feature extraction, \\ motion prediction, affinity computation, and association methods} & General & 2019 \\[0.5cm]
\citep{survey2020_associaton_model_based} & Model-based association methods in MOT & General & 2020 \\[0.5cm]
\citep{survey2021_autonomous_vehicles} & MOT methods for autonomous vehicles & Domain & 2021 \\[0.5cm]
\citep{survey2021_deep_learning_review_by_components} & \makecell[l]{Transition from traditional to deep learning-based MOT methods: \\ Emphasis on tracking-by-detection approaches} & General & 2021 \\[0.5cm]
\citep{survey2022_traffic_environmnets} & MOT methods in traffic environments & Domain & 2022 \\[0.5cm]
\citep{survey2022_association_methods} & Association methods in MOT: Successes and future challenges & General & 2022 \\[0.5cm]
\citep{survey2024_embedding_methods} & Embedding methods in MOT: A comprehensive review & General & 2022 \\[0.5cm]
\citep{mot_gnn_survey} & Graph-based data association in MOT & General & 2023 \\
\end{tabular}
\caption{A summary of literature reviews on Multi-Object Tracking (MOT).}
\label{tab:related_review_papers}
\end{table*}

\textbf{Our focus}. Our survey paper falls into the general surveys category. We focus on two main groups of works: pivotal studies that directly influenced modern trackers (e.g., SORT~\citep{sort} and Deep SORT~\citep{deep_sort}) and modern state-of-the-art deep learning-based trackers, beginning with tracking-by-detection approaches and extending to end-to-end trackers. For performance comparisons, we consider only high-performance trackers published from 2022 onward. This makes our work complementary to well-known MOT survey papers~\citep{survey2019_deep_learing_in_MOT, survey2021_deep_learning_review_by_components, survey2024_embedding_methods}, as they primarily cover older methods. Furthermore, to assess the generalizability of tracking methods across different domains, we define our own benchmark using weighted macro-averaged results across multiple datasets from diverse domains.

\section{Background}

We provide the foundational concepts necessary to follow this survey. First, we formally define the multi-object tracking (MOT) problem. Afterwards, we present an overview of the metrics used to evaluate tracker's performance. Finally, we introduce standard MOT benchmark datasets.

\subsection{Problem Formulation}
\label{sec:problem_formulation}

Multi-object tracking is performed on a video that captures a scene containing a set of objects of interest. Let $K$ be a set of identities of objects that are present in the video over a discrete time interval $\{1, 2, \dots, T\}$. At any time $t$, there exists a set of visible objects:
\begin{equation}
    \boldsymbol{X}_{t} = \{\boldsymbol{x}^{k}_t \mid k \in K_{t}\}
\end{equation}
where $\boldsymbol{x}^{k}_t$ is the ground truth\footnote{The ground truth annotation of an object, $\boldsymbol{x}^{k}_t$, includes annotated features. For 2D bounding box annotations, in addition to the identity, the ground truth provides the horizontal position ($x$), vertical position ($y$), width ($w$), height ($h$), and class ($c$).} of an object with identity $k$ at time $t$, and $\boldsymbol{K}_{t} \subseteq \boldsymbol{K}$ is a set of visible objects at time $t$. Objects may enter, exit, or even return to the scene, which leads to $\boldsymbol{K}_{t}$ varying over time. The objective is to compute the trajectories of all objects in the video. The trajectory of an object with identity $k$, visible at times $\boldsymbol{T}_k$, is defined as:
\begin{equation}
    \boldsymbol{\tau}^{k} = \{\boldsymbol{x}^{k}_t \mid t \in \boldsymbol{T}_k\}
\end{equation}
We note that $\boldsymbol{\tau}^{k}$ can be fragmented due to object being occluded or out of view during some parts of the video. The set of all video ground truth tracks is defined as:
\begin{equation}
\boldsymbol{\tau} = \{\boldsymbol{\tau}^{k} \mid k \in \boldsymbol{K}\}
\end{equation}
A MOT tracker is an algorithm that predicts video tracks $\hat{\boldsymbol{\tau}}$, aiming to closely match the ground truth tracks. In the next section, we provide an overview of standard metrics used to evaluate tracker performance by comparing the predicted $\hat{\boldsymbol{\tau}}$ with the ground truth $\boldsymbol{\tau}$. We denote $\hat{\boldsymbol{\tau}}^{k}$ as the \emph{predicted track} and $\hat{\boldsymbol{x}}^{k}_{t}$ as the \emph{predicted detection} (or just \emph{detection}).

\begin{figure*}
  \centering
  \includegraphics[width=0.9\linewidth]{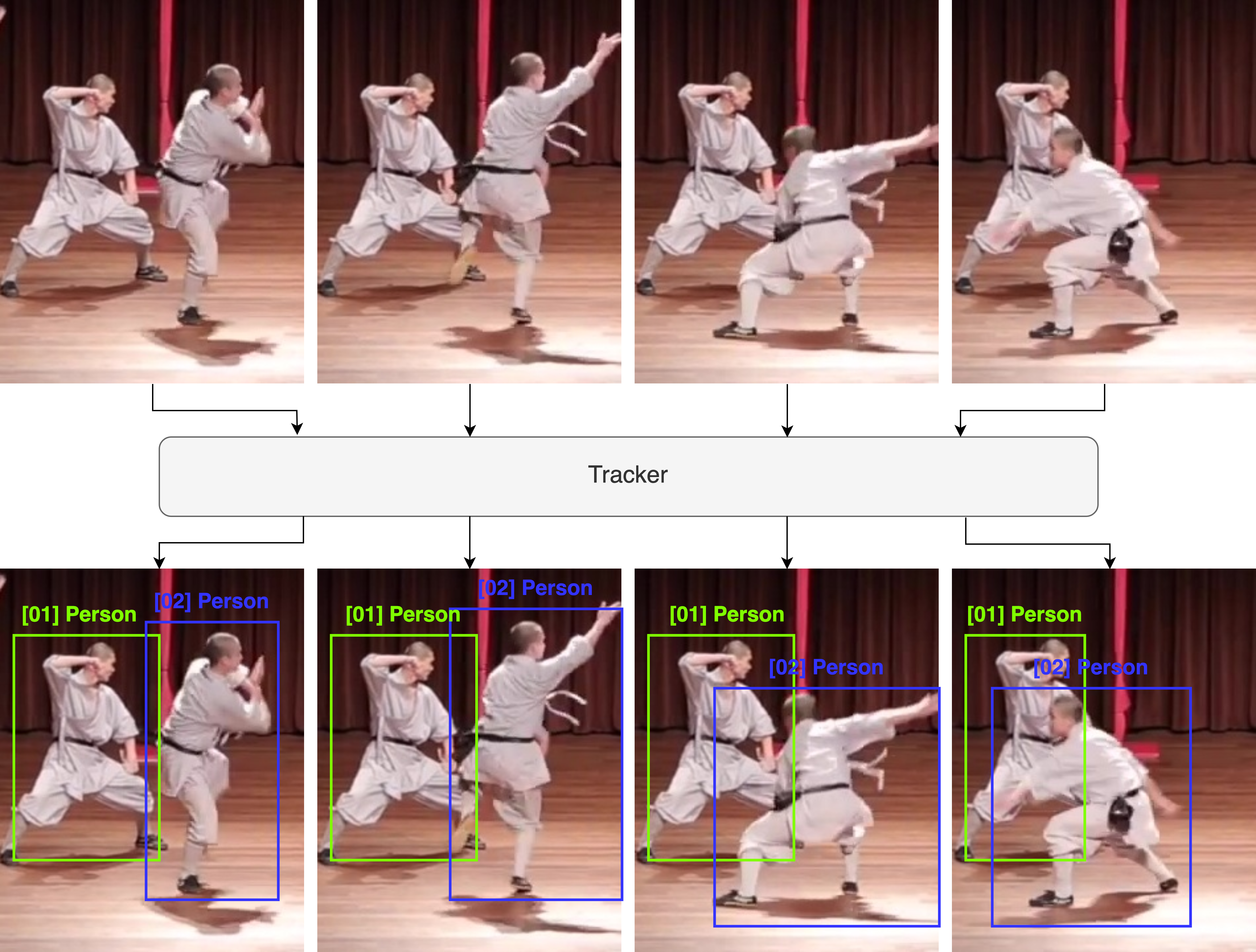}
  \caption{Illustration of a general MOT pipeline. Frames are taken from scene \textit{DanceTrack0001}~\citep{dancetrack}.}
  \label{fig:mot_illustration}
\end{figure*}

We illustrate the general MOT pipeline in Figure~\ref{fig:mot_illustration}. The input to the tracker is a sequence of raw video frames, and the output consists of sets of bounding boxes for each frame with their predicted coordinates, classes, and identities.

\subsection{Evaluation Metrics}

We describe the metrics used to compare predicted tracks, $\hat{\boldsymbol{\tau}}$, with ground truth tracks, $\boldsymbol{\tau}$. We begin by defining the error types, followed by the similarity measures and matching\footnote{We use the term \emph{match} to refer to pairing a prediction with a ground truth during evaluation, and \emph{association} to refer to pairing a track with a detection during inference.} between the ground truth and predictions. Finally, we provide an overview of commonly used metrics for evaluating tracking methods.

\textbf{Error types}. We define the following error types:

\begin{itemize} 
    \item $\emph{False positive track}$: A tracked object does not exist. 
    \item $\emph{False negative track}$: An object is not recognized throughout the video. 
    \item $\emph{Identity switch}$: Two or more tracks have mixed detection matches.    
    \item $\emph{False positive detection}$: An object's ground truth does not exist at a specific time (e.g., the object is out of view) but it was still falsely detected. 
    \item $\emph{False negative detection}$: An object's ground truth exists but is not detected at a specific time. 
    \item $\emph{Localization error}$: An object's detection offsets from the ground truth.
\end{itemize}

\begin{figure*}
  \centering
  \includegraphics[width=0.8\linewidth]{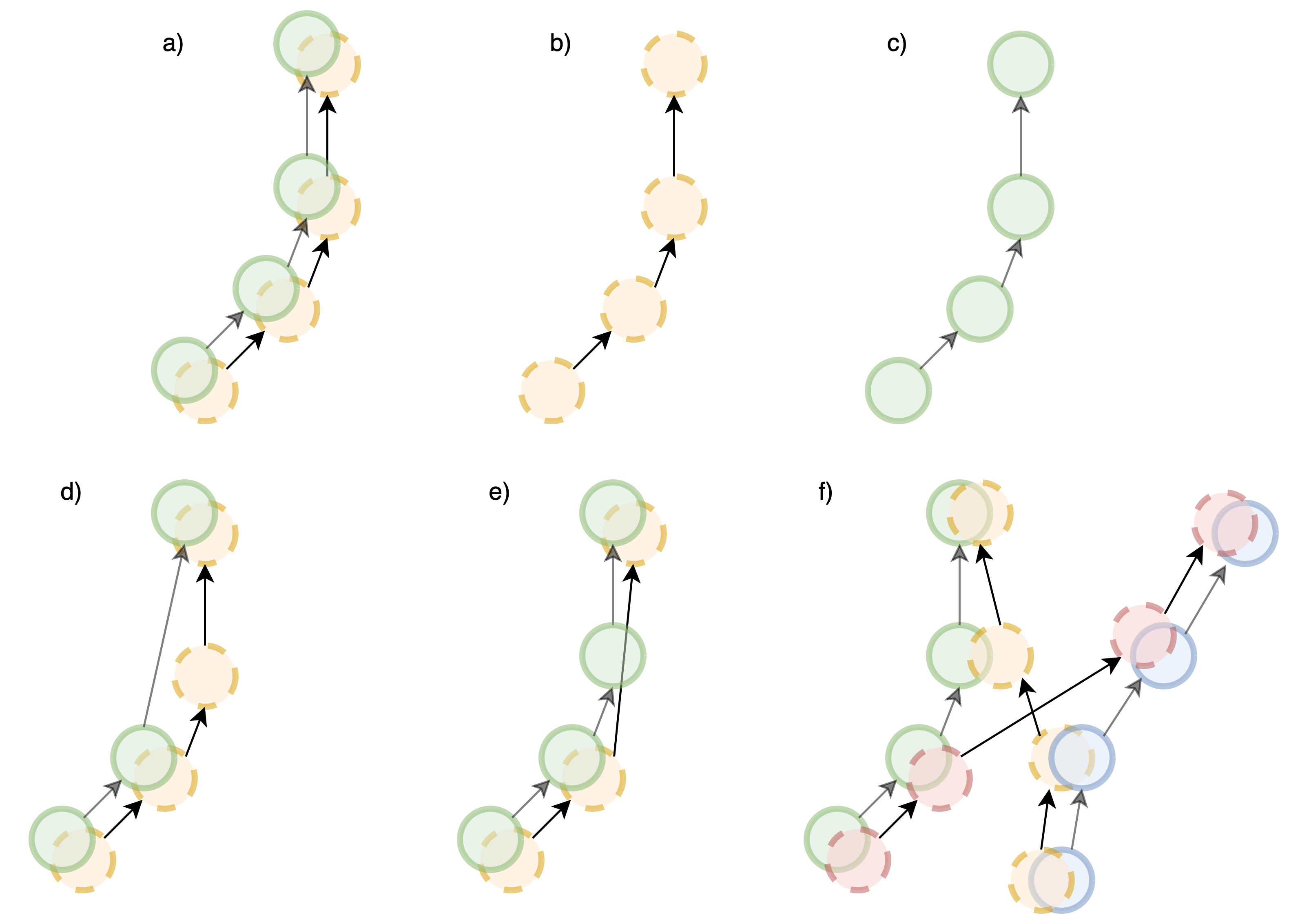}
  \caption{The figure illustrates six cases of tracking results. We differentiate between ground truth tracks, shown with ground truths as solid circles, and predicted tracks, shown with detections as dashed circles. In case (a), the predicted and ground truth tracks match at every moment. In case (b), we observe only the predicted track (false positive track). Similarly, in case (c), we see only the ground truth track, as the predicted track is missing (false negative track). In case (d), the predicted track misses the third detection (false negative detection), and (e) contains an extra, third detection (false positive detection). Finally, in case (f), both predicted tracks initially align but switch identities at the third step (identity switch). In all cases, we observe that the ground truths and predictions do not align perfectly, leading to localization errors.}
  \label{fig:error_types}
\end{figure*}

The first three types of errors are classified as \emph{association errors}, and last three are \emph{detection errors}~\citep{hota}. In Figure~\ref{fig:error_types}, we illustrate an example of correct tracking alongside examples of each previously mentioned error type. Localization errors are not explicitly visualized, as they may occur in all cases, given that predicted detections almost never perfectly match the ground truth.

\textbf{Similarity measure}. In Figure~\ref{fig:error_types}, we visually observe that the ground truths and predicted detections match up to some point. To qualitatively evaluate trackers, we first need to define a measure of similarity. The \emph{similarity score} should be chosen based on the domain, with possible values usually ranging from 0 to 1. For a prediction to match a ground truth, a similarity threshold must be met. If the objects' detections are represented as 2D or 3D bounding boxes, the standard IoU (Intersection over Union), also known as Jaccard index, can be used as the similarity measure. Alternatively, if detections are points (e.g., for pose estimation), a similarity measure based on Euclidean distance~\citep{hota} is more appropriate. In this paper, we focus on videos with objects detected as 2D bounding boxes.

\textbf{Bijective Matching}. During evaluation, multiple detections can overlap with a single ground truth, and vice versa. Effectively, all possible matching pairs form a bipartite graph, where each prediction belongs to the first group of nodes, and each ground truth belongs to the second group. The edges are weighted based on negative similarity (or positive distance) scores for minimal cost matching. A score threshold is often applied to determine whether two nodes can be matched; otherwise, the edge weight is treated as infinite. This setup represents a standard bipartite graph optimal assignment problem, where the objective is to maximize similarity or a metric of interest. It is commonly solved using the Hungarian algorithm~\citep{linear_assignment}. Matching is generally performed either at the detection level (i.e., per frame) or at the track level (i.e., across the video). In further text, we cover standard MOT metrics used to compare models on benchmarks.

\textbf{MOTA and MOTP metrics}. One of the earliest multi-object tracking metrics, which is still widely used today, is MOTA (Multi-Object Tracking Accuracy) along with MOTP (Multi-Object Tracking Precision)~\citep{mota, hota}. In this case, bijective matching is performed at the detection level, with default IoU threshold equal to $0.5$. Specifically, the MOTA metric is defined as follows:
\begin{equation}
    \text{MOTA} = 1 - \frac{|\text{FP}| + |\text{FN}| + |\text{IDSW}|}{|\text{gtDet}|}
\end{equation}
where $|\text{FP}|$ is the total number of false positive detections, $|\text{FN}|$ is the total number of false negative detections, 
$|\text{IDSW}|$ is the total number of identity switches, and $|\text{gtDet}|$ is the total number of ground truth detections. 
The MOTA metric values range from $-\infty$ to $1$, where higher values indicate better performance.

Since MOTA does not account for localization error, which measures average similarity over correct matches, MOTP is introduced and defined as:
\begin{equation}
    \text{MOTP} = \frac{1}{|\text{TP}|}\sum_{m \in \text{TP}} S_{m}
\end{equation}
where $|\text{TP}|$ is the total number of true positives (correct matches), and $S_m$ is the similarity score of the matched pair $m$. A higher MOTP indicates a higher average similarity score between matched ground truths and predictions. Matching is performed at the detection level by optimizing both MOTA and MOTP.

While MOTA and MOTP are useful metrics, relying solely on them is insufficient. Firstly, MOTA is more influenced by detection performance than association performance~\citep{mota, hota}. Additionally, it does not account for when an identity switch occurs---whether at the start, middle, or end of a track. For these reasons, other identification metrics are often used.

\textbf{IDF1 metric}. Opposite of the MOTA, the IDF1 metric calculates bijective matching at track level, which results in a different type of false positive and false negative detections. The IDF1 metric is defined as follows~\citep{mota, hota}:
\begin{equation}
    \text{IDF1} = \frac{|\text{IDTP}|}{|\text{IDTP}| + 0.5 \cdot |\text{IDFN}| + 0.5 \cdot |\text{IDFP}|}
\end{equation}
where $|\text{IDTP}|$ represents the number of true positive identity matches, $|\text{IDFN}|$ denotes the number of false negative identity matches, and $|\text{IDFP}|$ denotes the number of false positive identity matches. Matching is performed at the track level by optimizing the IDF1 metric. In contrast to the MOTA metric, the IDF1 metric is biased toward association performance and is sensitive to identity switches. To further evaluate tracking performance, ID-Recall (IDR) and ID-Precision (IDP) are also used:
\begin{equation} 
    \text{ID-Recall} = \frac{|\text{IDTP}|}{|\text{IDTP}| + |\text{IDFN}|} 
\end{equation} 
\begin{equation} 
    \text{ID-Precision} = \frac{|\text{IDTP}|}{|\text{IDTP}| + |\text{IDFP}|} 
\end{equation} 
where ID-Recall measures the percentage of true positive identity matches relative to the total number of ground truth matches (true positives and false negatives). ID-Precision measures the percentage of true positive identity matches relative to the total number of predicted matches (true positives and false positives). The IDF1 metric combines both IDP and IDR into a single metric, providing a balanced evaluation of identity matching performance~\citep{idf1}.

\textbf{HOTA metric}. HOTA (Higher Order Tracking Accuracy) is a metric designed to evaluate MOT performance by jointly 
considering detection, association, and localization quality. HOTA measures how well the tracks of matched detections align and 
averages this over all matched detections~\citep{hota}. 
Unlike earlier metrics such as MOTA and IDF1, which emphasize either detection or association, HOTA balances these aspects by explicitly separating 
detection and association errors while incorporating spatial and temporal consistency. For this metric, matching is performed at the detection level,
ensuring evaluation is based on individual object detections rather than higher-level track groupings.

\begin{figure*}
  \centering
  \includegraphics[width=0.4\linewidth]{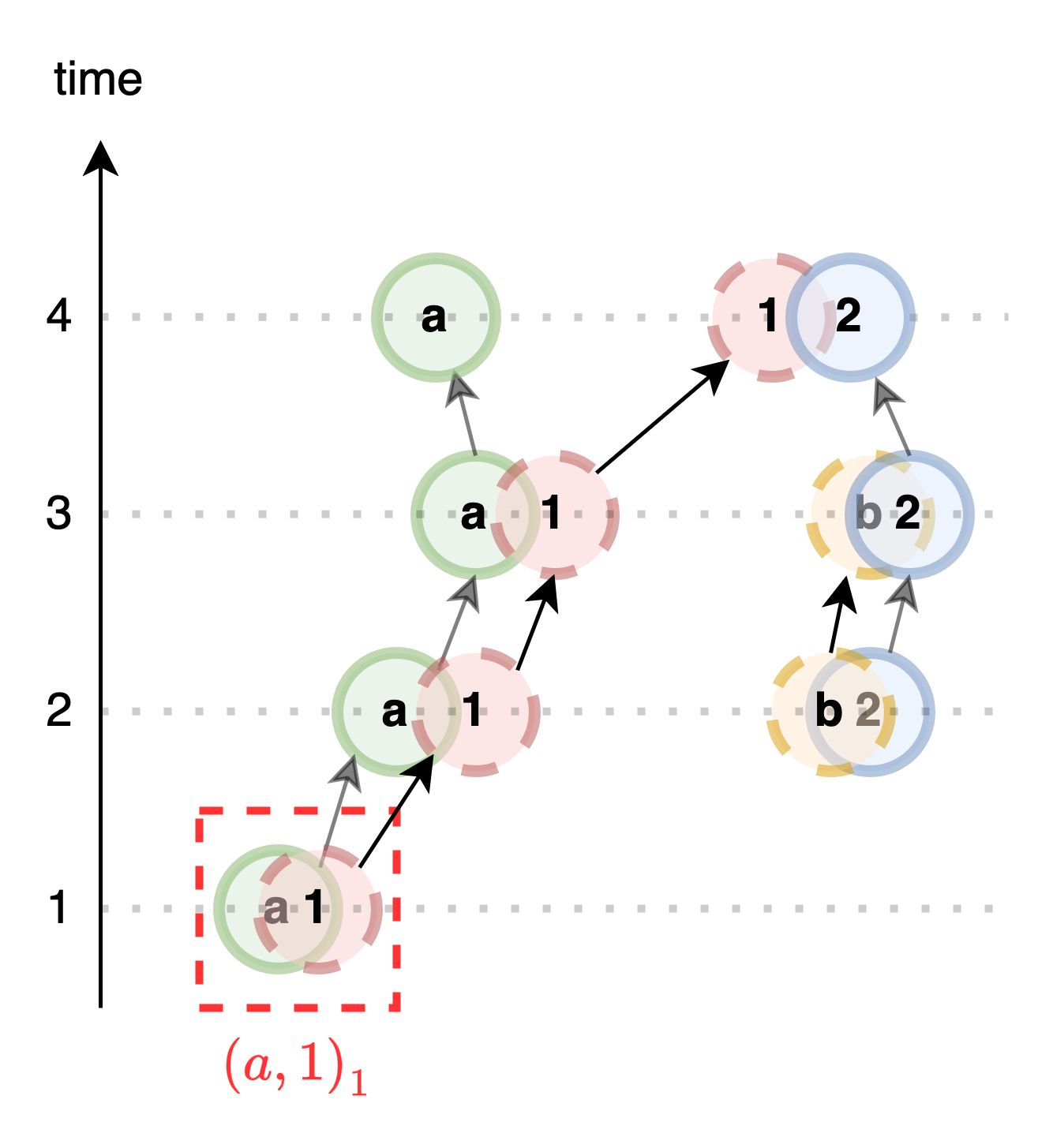}
  \caption{Example of ground truth and prediction track matching. The TP pair of interest is marked with a dashed red triangle. }
  \label{fig:assa_example}
\end{figure*}

The $\text{HOTA}_\alpha$ metric with a IoU similarity threshold $\alpha$ is defined as the geometric mean of detection $\text{DetA}_{\alpha}$ and association $\text{AssA}_{\alpha}$ metrics:
\begin{equation} 
    \text{HOTA}_{\alpha} = \sqrt{\text{DetA}_{\alpha} \cdot \text{AssA}_{\alpha}} 
\end{equation} 
The $\text{DetA}_{\alpha}$ metric is simply the Jaccard index, a measure of similarity between two sets --- ground truths and predictions. It is defined as:
\begin{equation} 
    \text{DetA}_{\alpha} = \frac{|\text{TP}|}{|\text{TP}| + |\text{FN}| + |\text{FP}|}
\end{equation} 
where TP is the set of matches between ground truths and predictions, FP is the set of predictions without a corresponding ground truth, and FN is the set of ground truths without a corresponding prediction. The $\text{AssA}$ metric is defined as:
\begin{equation} 
    \text{AssA}_{\alpha} = \frac{1}{|\text{TP}|} \sum_{c \in \{\text{TP}\}} \text{A}(c)
\end{equation} 
where
\begin{equation}
    \text{A}(c) = \frac{|\text{TPA}(c)|}{|\text{TPA}(c)| + |\text{FNA}(c)| + |\text{FPA}(c)|}
\end{equation}
In this case, $c$ is a pair of ground truth and prediction IDs. Specifically, set $\text{TPA}(c)$ is the set of true positives (TPs) that have the same ground truth and prediction IDs. Set $\text{FNA}(c)$ is the union of the set of all TPs that have the same ground truth ID as $c$ but a different prediction ID, and the set of all unmatched ground truths with the same ground truth ID as $c$. Set $\text{FPA}(c)$ is the union of the set of all TPs that have the same prediction ID as $c$ but a different ground truth ID, and the set of all unmatched predictions with the same ID as $c$~\citep{hota}. This is best understood through an example. Assume that we have the following set of matched pairs: $\{(a, 1)_{1}, (a, 1)_{2}, (b, 2)_{2}, (a, 1)_{3}, (b, 2)_{3}, (a, -)_{4}, (b, 1)_{4}\}$, where $(x, y)_{t}$ represents matched ground truth $x$ with prediction $y$ at time $t$. A visualization of this example can be seen in Figure~\ref{fig:assa_example}. 

In this case:
\begin{itemize}
    \item $TPA((a, 1)_{1}) = \{(a, 1)_{1}, (a, 1)_{2}, (a, 1)_{3}\}$, 
    \item $FNA((a, 1)_{1}) = \{(a, -)_{4}\}$,
    \item $FPA((a, 1)_{1}) = \{(-, 1)_{4}\}$,
    \item hence, $A((a, 1)_{1}) = \frac{3}{3 + 1 + 1} = 0.6$,
    \item similarly, $A((b, 2)_{2}) = \frac{2}{2 + 0 + 1} = 0.67$
\end{itemize}

Finally, in order to measure localization, HOTA is calculated with different thresholds ($0.05$ to $0.95$ in $0.05$ intervals):
\begin{equation}
    \text{HOTA} = \frac{1}{19} \sum_{\alpha} \text{HOTA}_{\alpha}
\end{equation}

Because HOTA considers both types of errors---association and detection---it is often used as the main metric for tracker benchmarks. Besides HOTA, for more detailed comparisons, AssA, DetA, MOTA, IDF1, and IDSW (total number of ID switches) are also evaluated~\citep{mot_challenge, dancetrack, sportsmot, kitti}. For all metrics except IDSW, higher values indicate better tracker performance.

\subsection{Datasets}

We provide an overview of the datasets commonly used as standard benchmarks for MOT methods. These datasets will later be used for method comparisons.

\textbf{MOTChallenge---MOT17 and MOT20}. The MOT17 dataset\footnote{MOTChallenge website: \url{https://motchallenge.net/}.} consists of 14 short video sequences captured with non-static cameras, predominantly featuring pedestrians with linear motion. These sequences are evenly split between training and test sets. Similarly, the MOT20 dataset includes eight heavily crowded scenes recorded with static cameras, also evenly divided between training and test sets. However, MOT20 contains approximately three times more bounding boxes in total compared to MOT17. The primary challenges in the MOTChallenge benchark's scenes are crowded environments and frequent occlusions~\citep{mot_challenge}.

\textbf{DanceTrack}. DanceTrack\footnote{DanceTrack GitHub page: \url{https://github.com/DanceTrack/DanceTrack}.} is a dataset comprising 100 videos of people dancing, each meticulously annotated for the purpose of MOT tracker design and evaluation. This task is particularly challenging due to the dancers' similar appearances, frequent occlusions, and their rapid, non-linear movements. The dataset is segmented into 40 training videos, 25 validation videos, and 35 test videos. The dataset is highly diverse as it includes multiple genres of dancing: classical, street, pop, large group and sports~\citep{dancetrack}. 

\textbf{SportsMOT}. SportsMOT\footnote{SportsMOT GitHub page: \url{https://github.com/MCG-NJU/SportsMOT}.} is a comprehensive multi-object tracking dataset that includes $240$ video sequences, totaling over $150,000$ frames. The dataset encompasses three categories of sports: basketball, volleyball, and football. It is particularly challenging for object tracking due to the fast and varying speeds of motion and the similar appearances of the objects involved. The dataset is organized into training, validation, and test sets, which contain $45$, $45$, and $150$ video sequences, respectively. Annotations include 2D bounding boxes, player identities, and team labels, enabling evaluations of trackers in complex scenarios involving frequent occlusions, rapid movements, and team-based associations~\citep{sportsmot}.

\textbf{Datasets overview}. Table~\ref{tab:dataset_stats} presents key statistics for five multi-object tracking datasets. MOT17 and MOT20 emphasize dense pedestrian tracking, with MOT20 containing significantly more bounding boxes (1.65M vs. 292K) and tracks (3456 vs. 1342) for the same total video length. On average, MOT20 has nearly five times as many pedestrians per frame as MOT17. When comparing datasets regarding the video length, DanceTrack and SportsMOT surpass MOTChallenge by an order of magnitude, featuring 105K and 150K frames, respectively, compared to just 27K frames for MOT17 and MOT20. However, we observe that these datasets are much less crowded than MOT20.

\begin{figure*}
  \centering
  \includegraphics[width=0.8\linewidth]{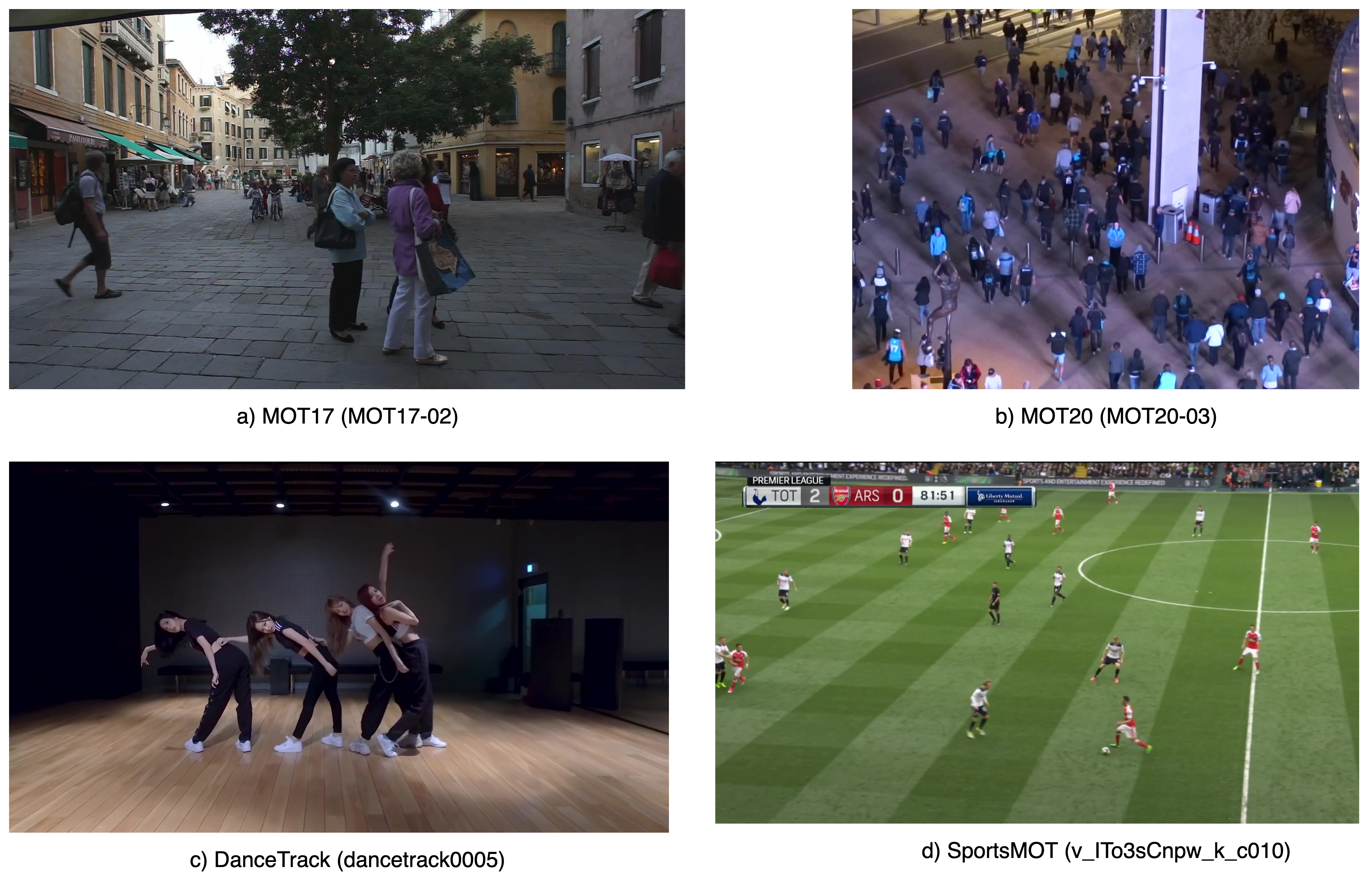}
  \caption{Sample frames from MOT datasets across different domains: (a) MOT17, (b) MOT20, (c) DanceTrack, (d) SportsMOT.}
  \label{fig:dataset_examples}
\end{figure*}

As it can be seen from frame samples in Figure~\ref{fig:dataset_examples}, these datasets collectively cover various challenges, including dense crowds, rapid motion, and long-term tracking. Evaluating a tracker on a single dataset is not ideal, as it fails to demonstrate generalization across different scenarios and motion patterns. MOTChallenge has objects with predictable motion with very crowded scenes. DanceTrack emphasizes diverse motion patterns and similar appearances. SportsMOT is a more extreme variant, featuring greater motion diversity and higher appearance similarity.

\begin{table*}
\small
\centering
\renewcommand{\arraystretch}{1.4}
\begin{tabular}{c|c c c c c}
Dataset & Scenes & Frames & Length (s) & BBox & Tracks \\
\hline
MOT17 & 14 & 14235 & 463 & 292733 & 1342 \\
MOT20 & 8 & 13410 & 535 & \textbf{1652040} & \textbf{3456} \\
DanceTrack & 100 & 105855 & 5292 & - & 990 \\
SportsMOT & \textbf{240} & \textbf{150379} & \textbf{6015} & 1629490 & 3401 \\
\end{tabular}
\caption{Summary of dataset statistics, including the number of scenes, frames, sequence length (in seconds), bounding boxes (BBox), and object tracks. Sources: ~\citep{mot_challenge, dancetrack, sportsmot}.}
\label{tab:dataset_stats}
\end{table*}

\section{Tracking-by-Detection Paradigm}
\label{sec:tbd}

Tracking-by-detection is a simple yet highly efficient paradigm in MOT, excelling in both accuracy and speed. The detection and association are decoupled and performed in independent steps. First, an object detection model detects objects (usually as 2D bounding boxes) in each frame. Then, an association method is used to associate objects across frames. This paradigm is widely adopted due to its modularity, which enables independent optimization of detection and association modules, its flexibility to integrate both traditional algorithms and deep learning models, and its strong performance in modern applications.

To make this section accessible, we begin by reviewing the literature on deep-learning-based object detection models. We then examine online tracking-by-detection methods, starting with the baselines---SORT and Deep SORT. Next, we present clustered MOT methods, covering joint detection and embedding, heuristic-based, motion-based, affinity learning, and offline (graph-based) approaches. Finally, we discuss the strengths, limitations, and future research directions of the tracking-by-detection paradigm. While our focus is on modern deep-learning-based methods, we also acknowledge a few older but highly influential MOT approaches.

\subsection{Deep learning object detection models}

Object detection models can be roughly categorized into three families: proposal-based methods, grid-based methods, and query-based methods~\citep{uoi}. We briefly introduce each of these families in further text.

\textbf{Proposal-based object detectors}. Proposal-based approaches follow a two-stage pipeline: first, region proposals are generated; then, each proposed region undergoes bounding box localization and classification. One of the earliest methods, R-CNN~\citep{rcnn}, generates region proposals (i.e., crops) using the selective search algorithm~\citep{selective_search} and then applies a convolutional neural network (CNN)~\citep{cnn, Goodfellow-et-al-2016} classifier for each proposal. Although R-CNN achieves high detection accuracy compared to other methods at the time, its main drawback is slow computation speed. Fast R-CNN~\citep{fast_rcnn} improves the efficiency of R-CNN by sharing convolutional features across proposals. First it uses a single CNN pass to obtain image features. Second, it introduces ROI (Region Of Interest) pooling, which pools image features for each proposal obtained through the selective search algorithm. This modification to the R-CNN results in a processing speedup, as the CNN extractor is applied once to the whole image instead of being applied separately to each crop. Fast R-CNN also introduces anchors—predefined bounding boxes of various sizes and aspect ratios—to improve detection accuracy by serving as reference points for object localization~\citep{fast_rcnn}. Further improvements were introduced in Faster R-CNN~\citep{faster_rcnn}, which replaces selective search with CNN-based region proposal networks (RPNs), enabling end-to-end training and significantly improving speed. Mask R-CNN~\citep{mask_rcnn} extends Faster R-CNN by adding a branch for pixel-wise instance segmentation, making it suitable for tasks requiring object masks in addition to bounding boxes. Both Faster R-CNN and Mask R-CNN are supported by the Detectron2 framework~\citep{detectron, detectron2}, a modular and extensible library commonly used to implement, train, and evaluate proposal-based object detectors.

The performance of these methods during inference often depends on Non-Maximum Suppression (NMS)~\citep{rcnn} to remove duplicate detections and retain the most confident predictions. This results in a two-step inference process, where object detection first outputs raw bounding boxes, potentially containing duplicates, and NMS is subsequently applied to eliminate most of them. NMS requires hyperparameter manual tuning based on the domain to avoid removing valid detections as duplicates.

\textbf{Grid-based object detectors}. YOLO (You Only Look Once)~\citep{yolo, yolo_survey} models are state-of-the-art in real-time performance~\citep{yolov10} with a single-shot design that directly predicts bounding boxes and class probabilities from an image. This object detection model uses a grid-based approach to detect objects in a single pass, dividing the image into a grid where each cell is responsible for detecting one or more objects. In this case, the object detection problem is formulated as a regression task, optimized using the well-known YOLO loss~\citep{yolo}. The main drawback of the original YOLO, compared to concurrent proposal-based methods, was its difficulty in detecting small objects due to the design of its regression loss. YOLOv2~\citep{yolov2} introduces a bag of tricks to improve performance in terms of both accuracy (e.g. better detection of smaller objects) and speed. Inspired by Fast R-CNN, it introduces anchors into the YOLO family, improving localization accuracy while maintaining real-time performance. YOLOv3~\citep{yolov3} enhances multi-scale predictions using feature pyramid networks (FPNs)~\citep{fpn}, further improving robustness for small objects. YOLOv4~\citep{yolov4} introduces an extended bag of tricks, including CSPDarkNet~\citep{cspnet} as a backbone and mosaic augmentation, achieving strong performance while being lightweight enough to train on a single NVIDIA 2080 Ti GPU---unlike its predecessors. Ultralytics' YOLOv5, YOLOv8 and YOLOv11~\citep{ultralytics_yolo}, further optimize the architecture for speed and accuracy, incorporating advanced techniques such as mosaic augmentation, auto-anchor optimization, and better backbone designs. YOLOv5 is the first YOLO model implemented in Python, specifically using PyTorch~\citep{pytorch}. YOLOX~\citep{yolox} adopts an anchor-free approach with decoupled heads for classification and regression, achieving strong performance in accuracy and tracking applications~\citep{dancetrack, sportsmot, bytetrack, deep_eiou}. YOLOX is often used for MOT benchmarks~\citep{bytetrack, dancetrack, sportsmot}. Ultralytics' YOLO models also tend to be used due to their speed and ease of integration~\citep{yolov5yolov8yolov10goto, yolov8review, yolov8shiptracker}. YOLO models also depend on NMS to remove duplicate detections. Although anchorless versions are less reliant on NMS, they still require its use for optimal performance~\citep{yolox, ultralytics_yolo}.

\textbf{Query-based object detectors}. DETR (DEtection TRansformer)~\citep{detr} reformulates object detection as a set prediction problem, eliminating the need for hand-crafted anchors and post-processing steps such as the NMS, enabling both end-to-end training and inference. The model consists of four main components: a CNN backbone (typically ResNet~\citep{resnet}) that extracts image features, an transformer~\citep{attention_is_all_you_need} encoder that further enriches these image features using self-attention, a decoder that predicts object embeddings based on learnable queries (learnable positional encodings) through cross-attention with image features, and a prediction head that maps these object embeddings to bounding box coordinates and class labels. DETR's training process leverages a bipartite matching strategy to assign ground truth objects to predictions. Using the Hungarian algorithm, the model ensures a unique one-to-one assignment, simplifying the pipeline and enabling it to learn to avoid duplicate predictions---thus eliminating the need for NMS. Compared to the proposal-based and YOLO families, DETR is fully end-to-end, as it requires no NMS postprocessing, which can significantly impact CPU or GPU compute~\citep{rt_detr}. However, despite these advantages, DETR suffers from slow convergence, particularly for small objects.

DAB-DETR~\citep{dab_detr} improves upon DETR by introducing dynamic anchor boxes, addressing convergence speed and small object detection issues. DINO~\citep{dino_detr} builds on DETR with advanced query initialization and contrastive denoising, achieving state-of-the-art results in detection accuracy and robustness. However, DETR models are in general slow compared to YOLO models, unless optimized variants for real-time are used, like RT-DETR~\citep{rt_detr}. Besides object detection, DETR enabled a family of transformer-based end-to-end tracking methods which we will go through in detail in next section~\citep{motr, motrv2, motrv3, memot, memotr, motip}.

\subsection{SORT}

SORT (Simple online and real-time tracking)~\citep{sort} is the pioneer of tracking-by-detection paradigm. It's a simple algorithm with three key components: object detection model, association method and track handling. A motion model or an appearance model is often used to improve association performance.

\textbf{Object detection model}. For object detection, original SORT uses Faster R-CNN. This model was state-of-the-art at the time~\citep{faster_rcnn, sort}. However, SORT and other tracking-by-detection methods are not tied to a specific object detection model. Faster R-CNN can be easily replaced with any modern, more efficient model (e.g. YOLOX)~\citep{bytetrack}. 

Tracker only considers bounding boxes with high detection confidence. An important hyperparameter for tracker performance is the \emph{detection score threshold}, $det_{\tau}$, which balances false positives and false negatives. A higher threshold reduces false positives but increases false negatives, and vice versa. SORT’s performance is highly sensitive to the choice of $det_{\tau}$~\citep{bytetrack}.

\textbf{Motion model}. SORT uses a Kalman Filter (KF) estimation model to perform both bounding box motion prediction and noise filtering. A brief KF introduction can be found in~\ref{appendix:kalman_filter}. Particularly, SORT employs linear constant velocity model for track's motion prediction. This motion prediction is used during the association step to associate the track with detections. After a successful association, Bayesian inference is applied to combine information from both sources---the motion prediction and the associated detection---to update the track's state. The KF's state is modeled as $[x, y, s, r, \dot{x}, \dot{y}, \dot{s}]$, where $x$ and $y$ represent the bounding box center coordinates, $s$ is the bounding box scale (area), $r$ is the bounding box aspect ratio between height and width, $\dot{x}$ and $\dot{y}$ are the velocities of the center coordinates, and $\dot{s}$ is the scale velocity. Notably, this model assumes that the bounding box aspect ratio does not change over time. In context of Bayesian inference, the motion model prediction acts as the \emph{prior}, while the detection serves as the \emph{measurement likelihood} mean. Since object detectors typically do not provide uncertainty estimates for individual coordinates, heuristics are often used to approximate the measurement likelihood variance~\citep{sort, deep_sort, botsort}.

\textbf{Association method}. For each frame, SORT uses Hungarian algorithm to solve linear assignment problem in assigning detections to tracks. Association cost between a track's prediction bounding box and a detection bounding box is equal to negative IoU (Intersection Over Union). If the IoU similarity is below threshold $\text{IoU}_{min}$, the track cannot be associated with the corresponding detection, as they are too far apart. This type of constraints in assignment problems are often referred to as \emph{gates}. The input to a linear assignment solver can be represented as an association cost matrix, where rows correspond to tracks and columns correspond to detections. Linear assignment solver returns three sets as outputs: a set of associated tracks and detections pairs, a set of un-associated tracks, and a set of un-associated detections. Following the association step, the \emph{track handling} logic is responsible for managing all these sets.
    
\textbf{Track handling logic}. Tracks' creation and deletion should follow objects that are entering, leaving, and possibly returning to the scene over time. All un-associated detections are considered as potential new tracks with new identities. Creating a new track for each new detection would result in a high false positive ratio, for each object detection false positive. Instead, each track needs to undergo a probationary period, where it needs to be associated with a detection few frames in a row. If at least one frame is missed during the probationary period, the potential new track is deleted. The default probationary period in SORT is 3 frames~\citep{sort}. Existing tracks are terminated if they are not associated with any detection for $T_{lost}$ consecutive frames. Remembering the tracks is important in order to handle detector false negatives and objects' occlusions. SORT uses $T_{lost} = 1$ as a default configuration. This means that, with its default configuration, SORT does not handle long-range occlusions.

\textbf{Algorithm}. We define the following track states:  
\begin{itemize}  
\item \emph{active} — A state for objects that are currently being tracked with a successful association in the previous frame.  
\item \emph{lost} — A state for objects that are currently being tracked but failed to associate in one or more of the most recent frames (e.g. due to occlusion). Once  
      a lost track is successfully associated with a detection, it becomes active again.  
\item \emph{deleted} — A state for objects that are no longer being tracked and are considered completely lost.  
\item \emph{new} — A state for objects in a probationary period (potential tracks). Once the probationary period ends, the track becomes active.  
\end{itemize}  

These states are important in order to understand the implementation of SORT's Algorithm~\ref{alg:sort}. The three key steps of the algorithm are noted as: object detection inference, association and track handling. The motion model inference can be considered part of the association process. We omitted the probationary period logic for new tracks to keep the algorithm simpler. For the probationary period,  some changes are required in the track handling logic. Specifically, these tracks have to be handled during the \emph{update associated tracks} and \emph{update un-associated tracks} steps. 

\begin{algorithm}
\caption{SORT's pseudocode}
\label{alg:sort}

\hspace*{\algorithmicindent} \textbf{Models:} object detector \emph{Det}, Kalman Filter \emph{KF}; \\ 
\hspace*{\algorithmicindent} \textbf{Hyper-parameters:} association threshold $IoU_{min}$, detection threshold $det_{\tau}$, maximum track lost time $T_{lost}$; \\
\hspace*{\algorithmicindent} \textbf{Other:} Hungarian Algorithm \emph{HG} returns associated pairs, and un-associated singles; \\ 
\hspace*{\algorithmicindent} \textbf{Input:} a video sequence $\vee$; \\
\hspace*{\algorithmicindent} \textbf{Output:} predicted tracks $\hat{\tau}$ of the video.
\begin{algorithmic}[1]

\State \textbf{Initialization:} $\hat{\tau} = [\;]$
\For{\text{frame} $(t, f_{t})$ \textbf{in} V}
    \State \textbf{\# Object detection inference}
    \State $D = Det(f_{t}, det_{\tau})$ \\
    \State \textbf{\# Association}
    \State $\hat{T} = [\;]$  \Comment{Tracks' prior predictions}
    \For{$\tau_{k}$ \textbf{in} $get\_alive(\hat{\tau})$}  \Comment{alive --- active or lost}
        \State $\hat{T}.append(KF.predict(\tau_{k}, t))$
    \EndFor
    \State $C = -IoU(D, \hat{T}, IoU_{min})$
    \State $A, T_{unassoc}, D_{unassoc} = HG(C)$ \Comment{A --- Associated pairs} \\
    \State \textbf{\# Track handling}
    \For{$(i_{t}, i_{d})$ \textbf{in} $A$}  \Comment{Update associated tracks}
        \State $\hat{\tau}^{k_{i_{t}}} = \hat{\tau}[i_{t}]$
        \State $\hat{\tau}^{k_{i_{t}}}.update(KF.update(\hat{T}[i_{t}], D[i_{d}]))$
        \State $\hat{\tau}^{k_{i_{t}}}.lost = 0$
    \EndFor
    \For{$i_{t}$ \textbf{in} $T_{unassoc}$}. \Comment{Update un-associated tracks}
        \State $\hat{\tau}^{k_{i_{t}}} = \hat{\tau}[i_{t}]$
        \State $\hat{\tau}^{k_{i_{t}}}.lost = \hat{\tau}^{k_{i_{t}}}.lost + 1$
        \If{$\hat{\tau}^{k_{i_{t}}}.lost \geq T_{lost}$}
            \State $\hat{\tau}.delete(k_{i_{t}})$
        \EndIf
    \EndFor
    \For{$i_{d}$ \textbf{in} $D_{unassoc}$}. \Comment{Create new tracks}
        \State $d^{i_{t}} = d[i_{t}]$
        \State $\hat{\tau}.create(d^{i_{t}})$
    \EndFor
\EndFor

\end{algorithmic}
\end{algorithm}

\subsection{Deep SORT} 

SORT generally exhibits a high frequency of identity switches~\citep{sort, deep_sort}, primarily due to the low precision of its motion model. This issue is especially pronounced during long occlusions, where prediction errors accumulate.
Since SORT's motion-based association is insufficient in the general case, Deep SORT~\citep{deep_sort} proposed to combine the motion and appearance information for more accurate associations. 
Appearance information is extracted from images based on objects' bounding box crops (pathes). This is achieved using a convolutional neural network (CNN)~\citep{Goodfellow-et-al-2016} trained on a re-identification dataset in classification mode to discriminate between objects' identities~\citep{mars_reid}. 
Specifically, a trained CNN extracts each detected object's appearance embedding vector, which then passes through a softmax classification head to output the object's identity. During inference, classification head is dropped and cosine distance is used to compare feature embedding vectors, yielding a similarity score between $0$ and $1$. A model used for appearance embedding extraction is often called a \emph{re-ID} model, as it is specifically trained for this purpose. 

\begin{figure*}
  \centering
  \includegraphics[width=0.8\linewidth]{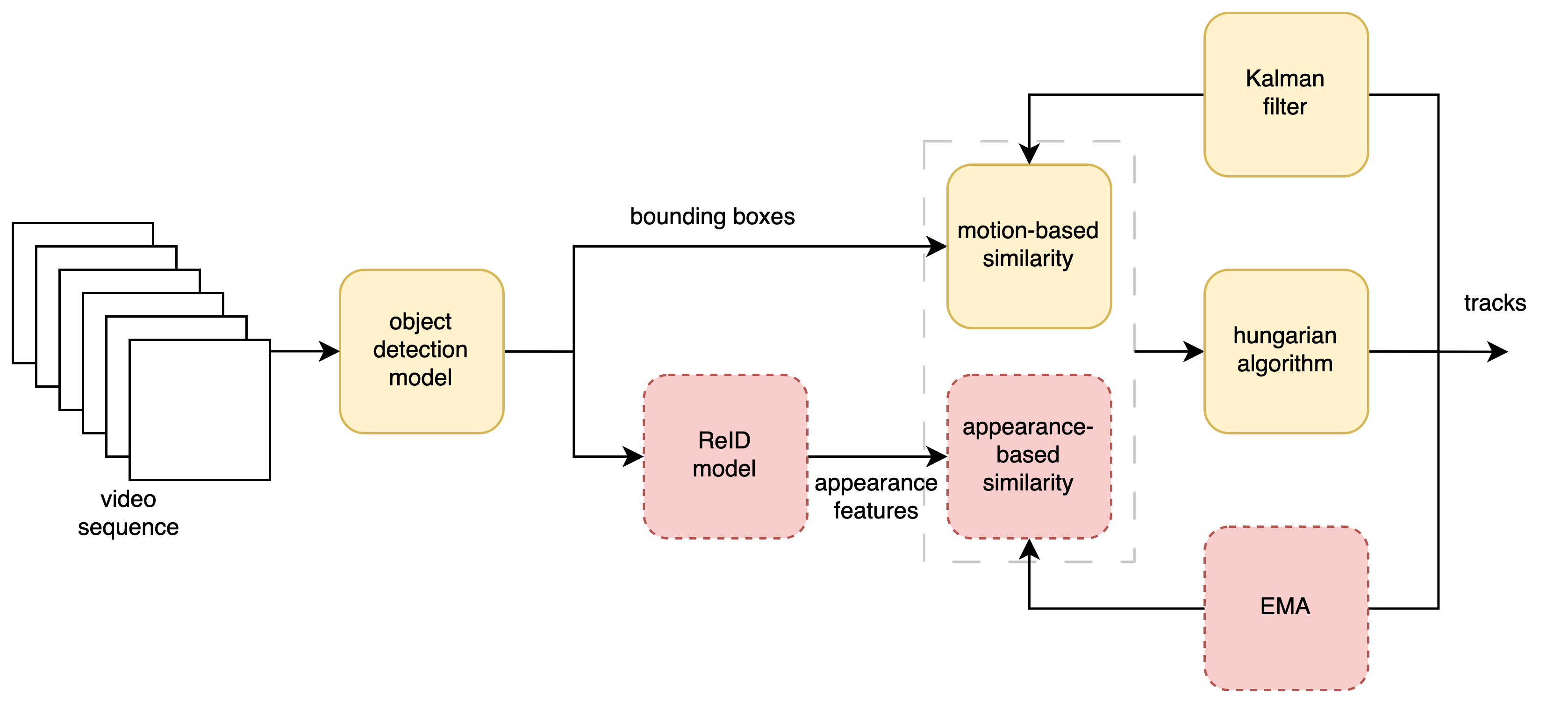}
  \caption{Overview of the DeepSORT tracking pipeline. Components shown in yellow with solid borders represent the original SORT framework, while components in red with dashed borders denote DeepSORT extensions. An object detection model generates bounding boxes, which are passed through a ReID model to extract appearance features. Similarity is then computed between the appearance features of the detections and those of existing tracks, maintained using a buffered history.}
  \label{fig:deepsort_pipeline}
\end{figure*}

DeepSORT's high-level tracking pipeline is shown in Figure~\ref{fig:deepsort_pipeline}. The detector and the re-ID model operate sequentially. First, the detector outputs a list of bounding boxes. Then, for each bounding box, the re-ID model extracts an appearance embedding, which is used for appearance-based association. Specifically, each track maintains a buffer of appearance feature embeddings. The similarity between a track and a detection is computed as the nearest neighbor distance between the detection's embedding and any embedding in the track's buffer.
\begin{equation}
    d_{A}(i, j, \boldsymbol{E}^{D}, \boldsymbol{E}^{T}) = min_{\boldsymbol{E}^{T}_{j,k} \in \boldsymbol{E}^{T}_{j}}d_{cos}(\boldsymbol{E}^{D}_{i}, \boldsymbol{E}^{T}_{j,k})\label{eq:deep_sort_appearance_cost}
\end{equation}
where $d_{cos}$ is the cosine distance, $i$ is the index of the $i$th detection, $j$ is the index of the $j$th track, $\boldsymbol{E}^{D}$ represents the detection appearance embeddings, and $\boldsymbol{E}^{T}$ represents the buffered track appearance embeddings, in which case $\boldsymbol{E}^{T}_{j,k}$ is the $k$th embedding in the $j$th track's embedding buffer.

In addition to integrating appearance information into association, Deep SORT introduces several other modifications. 
First, for the KF state it uses height and width along with their velocity states instead of scale and aspect ratio. 
This change provides greater flexibility to the KF model, as height and width have decoupled velocities. 
Second, the cost function for the assignment problem is modified. 
Instead of relying on the negative IoU, Deep SORT employs the Mahalanobis distance, which accounts for the uncertainty provided by the KF's predictions. 
The cost function is then defined as:  
\begin{equation}
    C_{\text{Deep SORT}}(i, j, \boldsymbol{D}, \boldsymbol{T}, \boldsymbol{E}^{D}, \boldsymbol{E}^{T}) = \lambda d_{M}(\boldsymbol{D}_i, \hat{\boldsymbol{T}}_j) + (1 - \lambda) d_{A}(i, j, \boldsymbol{E}^{D}, \boldsymbol{E}^{T}), \label{eq:deep_sort_cost}
\end{equation}
where $\boldsymbol{D}_i$ represents the $i$th detection with corresponding appearance embedding $\boldsymbol{E}_i$, $\hat{\boldsymbol{T}}_j$ represents the $j$th predicted track bounding box, $d_{M}$ is the Mahalanobis distance, and $d_{A}$ is the appearance distance. The weight $\lambda$ balances motion-based and appearance-based association costs. If $\lambda = 1$, the association cost is determined solely by motion and geometric features, whereas if $\lambda = 0$, it is based only on appearance features. Deep SORT uses $\lambda = 0$ as its default configuration~\citep{deep_sort}. However, motion and geometric features are still used in the gate function calculation, which is defined as follows:

\begin{equation}
    G_{\text{Deep SORT}}(i, j, \boldsymbol{D}, \boldsymbol{T}, \boldsymbol{E}^{D}, \boldsymbol{E}^{T}) = \boldsymbol{1}[d_{M}(\boldsymbol{D}_i, \hat{\boldsymbol{T}}_j) < t^{(M)}] \cdot \boldsymbol{1}[d_{A}(i, j, \boldsymbol{E}^{D}, \boldsymbol{E}^{T}) < t^{(A)}]\label{eq:deep_sort_gate}
\end{equation}
where $t^{(M)}$ is the Mahalanobis distance threshold and $t^{(A)}$ is the appearance distance threshold. Association is possible only if both motion and appearance conditions are met.

It is worth noting that the Mahalanobis distance is not a popular choice in the modern MOT algorithms due to its limitations~\citep{bytetrack, ocsort, botsort, sparsetrack, deep_eiou, deepocsort}. 
The problem lies in the fact that predictions for lost tracks accumulate uncertainty, which is natural but makes lost tracks cheaper to associate with compared to active tracks. 
To address this, Deep SORT introduced cascaded association based on track's time lost which partly resolves this issues. Empirical evidence suggests that this modification becomes less effective as the tracker performance improves~\citep{strongsort}.

The final significant modification is the decoupled association between active and new tracks, where detections are first associated with active tracks.  
This prioritization of already tracked objects over potential false positives further improves robustness.  
This modification is applied to modern MOT methods~\citep{bytetrack, botsort, sparsetrack, boostrack, ocsort, deepocsort, deep_eiou}.

In summary, the pipeline of Deep SORT, with its improved association performance, differs slightly from that of SORT. It introduces an additional appearance extraction step between detection inference and association, which can also be viewed as a sub-step of the association process. The pipeline remains decoupled, allowing the re-ID model to be improved or replaced without modifying any other component of the tracker~\citep{deep_sort}.

\subsection{Tracking-by-detection method categorization}

\begin{figure*}
  \centering
  \includegraphics[width=1.0\linewidth]{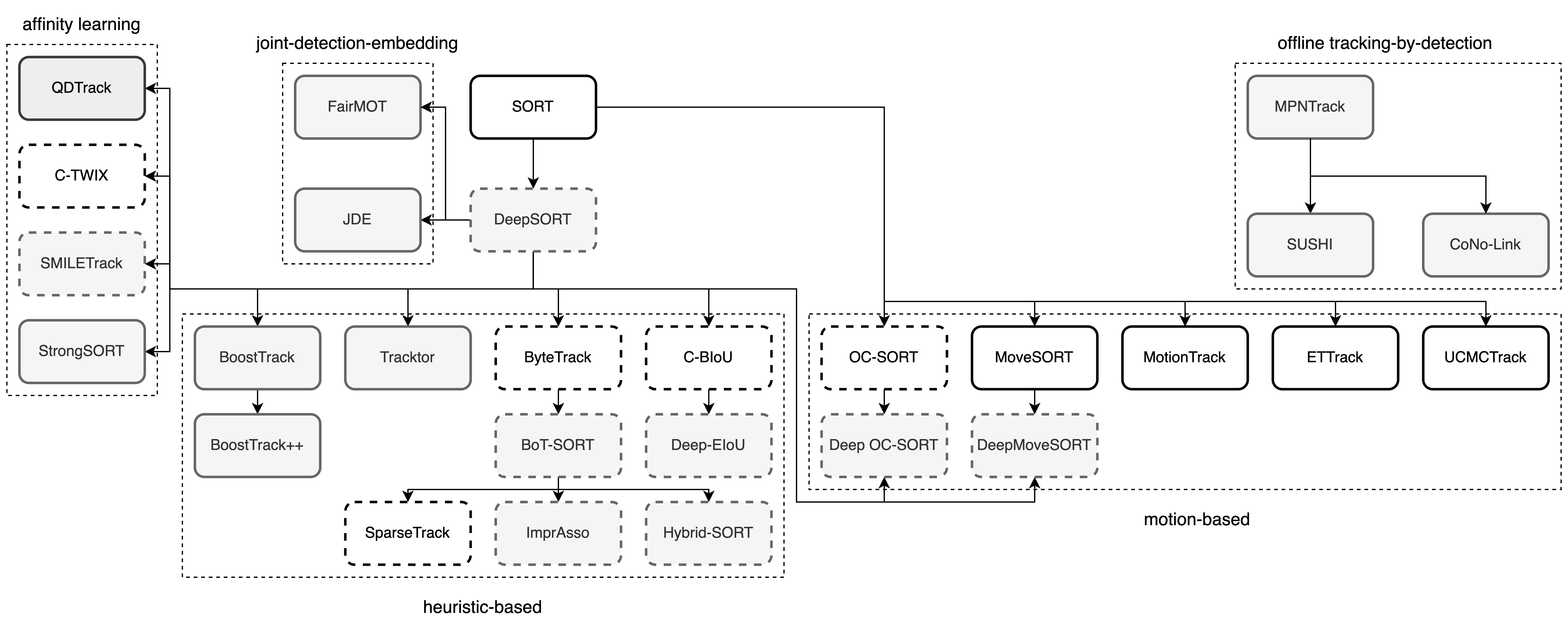}
  \caption{Evolution of influential tracking-by-detection Methods in MOT. Trackers with shaded boxes employ re-ID model, while trackers with dashed box borders use cascaded association (e.g. Byte).}
  \label{fig:tbd_research_advancements}
\end{figure*}

In the following text, we highlight the most important tracking-by-detection methods. Notably, every tracking-by-detection algorithm in MOT can be regarded as a variant of SORT. A graph illustrating the most influential work with emphasis on the recent studies is presented in Figure~\ref{fig:tbd_research_advancements}. These algorithms are roughly categorized into the following categories:  
\begin{itemize}
    \item \textit{Joint-detection-embedding} (JDE) methods, which propose a single model that performs both detection and appearance extraction, instead of using separate detection and re-ID models like in Deep SORT.  
    \item \textit{Heuristic-based} methods, which rely solely on association heuristics without introducing any model-related improvements.  
    \item \textit{Motion-based} methods, which enhance association accuracy by directly improving the motion model component.  
    \item \textit{Affinity learning} methods, which learn object affinities based on geometric, motion, or appearance features.  
    \item \textit{Offline} methods, which perform association at a global level after all video detections have been obtained.  
\end{itemize}  

It is important to note that most tracking-by-detection algorithms incorporate some form of heuristic to improve association accuracy. However, we refer to heuristic-based methods as those that do not propose any significant improvements (e.g., model enhancements) beyond a set of heuristics for improved association. In the following sections, we briefly review each of these algorithm categories, highlighting their progress over their predecessors.

\subsection{Joint detection and embedding tracking methods}

\begin{figure*}
  \centering
  \includegraphics[width=0.6\linewidth]{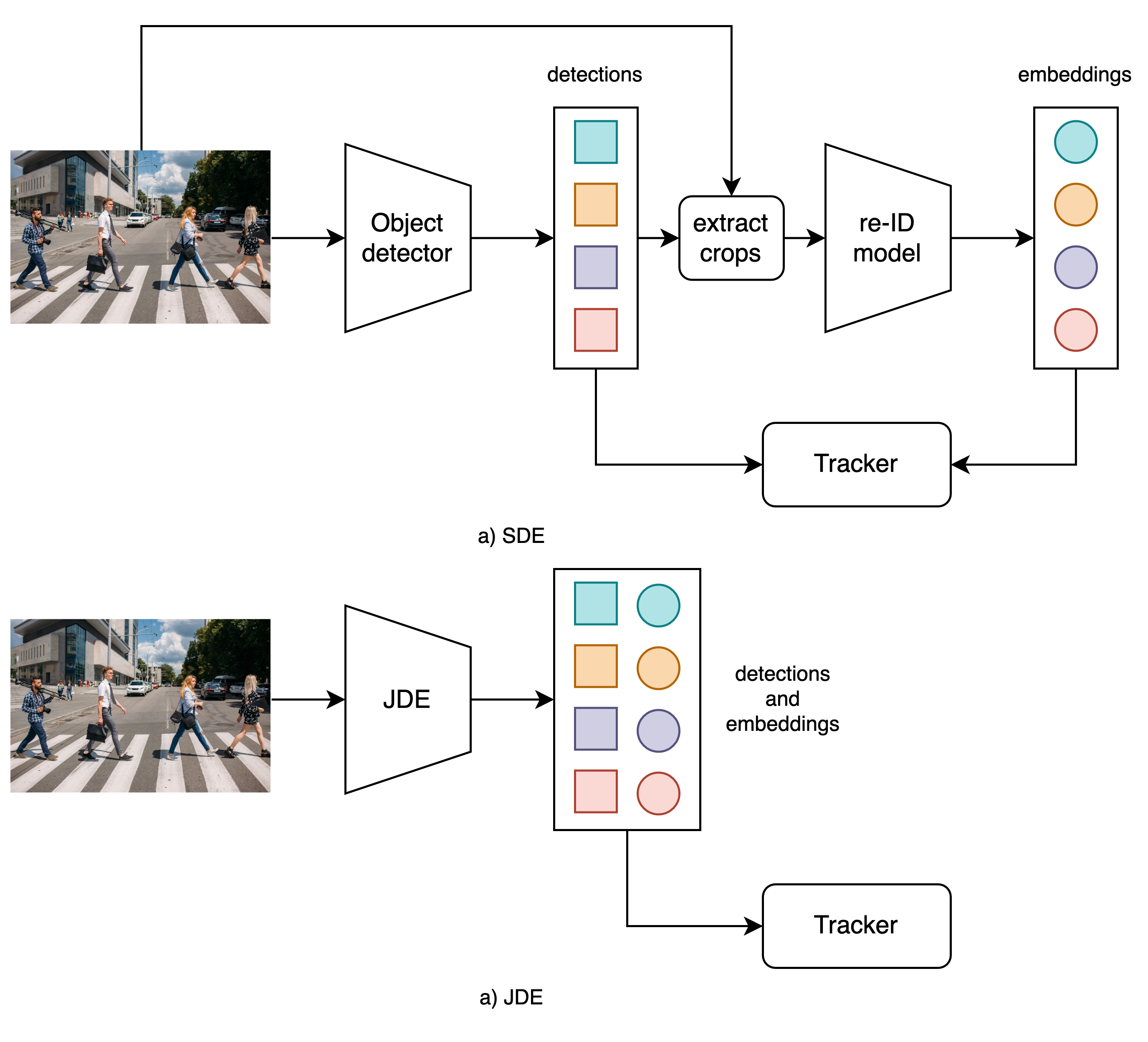}
  \caption{Architectural comparison between joint detection-embedding (JDE) and separate detection-embedding (SDE) methods.}
  \label{fig:jde_vs_sde}
\end{figure*}

\textit{JDE}. JDE (Joint Detection and Embedding)~\citep{jde} integrates detection and re-identification (re-id) tasks into a single neural network through joint task learning. The primary motivation behind this approach is to mitigate the slower inference of re-ID models in crowded scenes, where computational requirements scale with the number of cropped regions. By integrating the re-ID model within the detector, both tasks can be performed in a \emph{one-shot} manner, as shown in Figure~\ref{fig:jde_vs_sde}. 

Instead of maintaining a buffer for a track's historical appearance features, JDE employs an moving average over history of track's appearance features. The track's appearance update is as follows:
\begin{equation}
    \boldsymbol{E}^{T}_{j, f+1} = (1-\alpha)\cdot \boldsymbol{E}^{D}_{j, f+1} + \alpha \cdot \boldsymbol{E}^{T}_{j, f} \label{eq:jde_appearance_cost}
\end{equation}
where $\alpha$ is the moving average momentum, and $f$ is the index of the $f$th frame. These modifications improve tracking speed and robustness~\citep{jde}. Compared to Deep SORT, this tracking method achieves approximately four times higher FPS than Deep SORT variants, making it more suitable for real-time applications. However, its tracking accuracy is lower compared to these slower models.

JDE solves the re-ID task using metric learning instead of the classification approach employed in Deep SORT~\citep{jde, deep_sort}. In the classification approach, a re-ID model is trained to assign detections to specific identities present in the dataset. In contrast, metric learning maps detections into an embedding space, where the similarity between embeddings indicates the likelihood that a track and a detection belong to the same object identity. Metric learning generally outperforms classification when used as a standalone as it is better aligned with the requirements of MOT. However, hybrid approaches that combine both metric learning and classification tend to perform very well~\citep{fastreid_bot, botsort}. 

\textit{FairMOT}. One major drawback of the JDE tracker is the imbalance between the re-ID and detection tasks during training, in which case the features of these tasks often fail to align. Another issue stems from architectural design, as certain priors used for object detection (e.g., anchors and feature dimensions) can introduce challenges for re-id. FairMOT~\citep{fairmot} addresses these limitations by introducing key architectural changes, including an anchor-free CenterNet~\citep{centernet} detector and lower-dimensional re-ID features, thereby achieving a better balance between the two tasks. These changes led to significant improvements in tracker performance~\citep{fairmot}.

\textit{TransTrack}. We classify TransTrack as a unique method that belongs to neither the tracking-by-detection nor the end-to-end paradigm but instead sits between them. Among tracking-by-detection methods, it is most similar to JDE and FairMOT. 

\begin{figure*}
  \centering
  \includegraphics[width=0.8\linewidth]{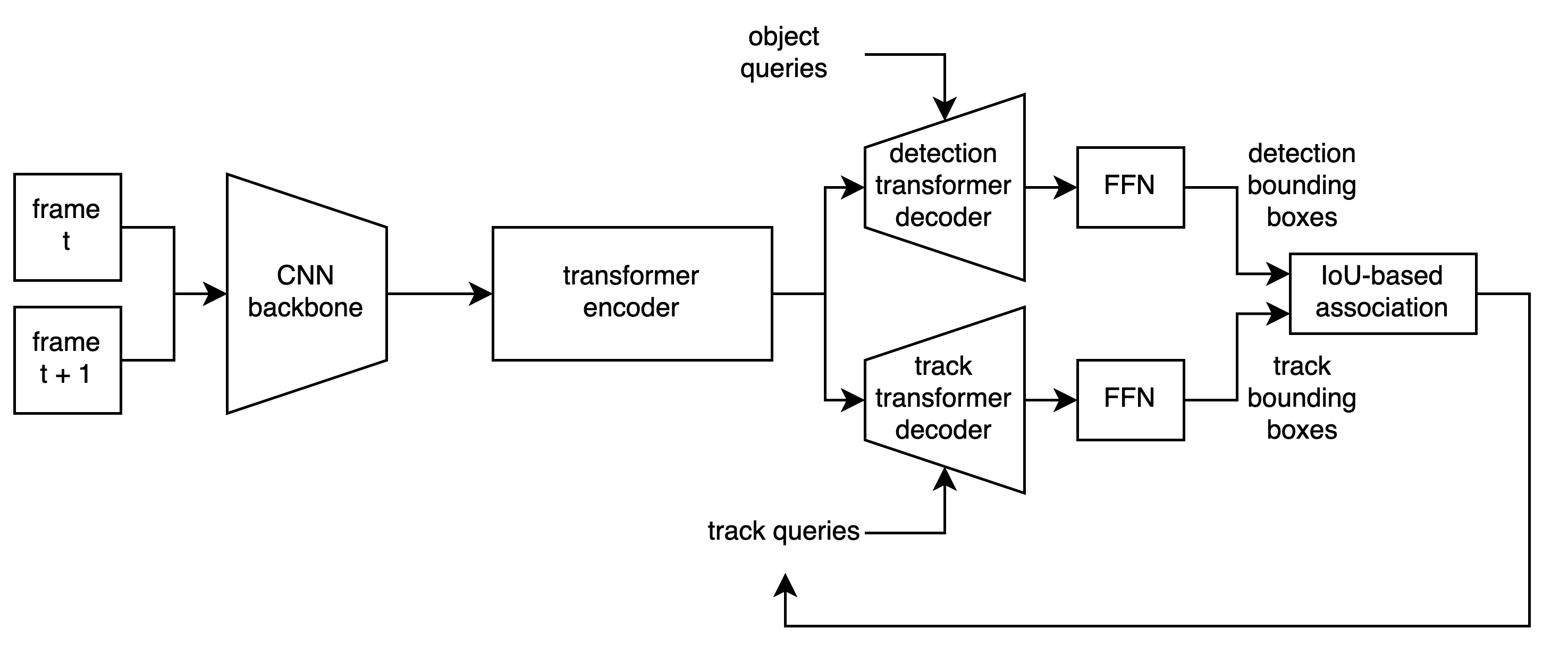}
  \caption{TransTrack architectural design.}
  \label{fig:transtrack}
\end{figure*}

TransTrack~\citep{transtrack} introduces a transformer-based approach that extends the DETR~\citep{detr} object detection model by adding an additional decoder specialized in track bounding box prediction. This decoder operates in parallel with DETR's standard detection decoder, as shown in Figure~\ref{fig:transtrack}. TransTrack forms a track query based on the track's history and combined features from the previous and current frames. This track query is analogous to the detection query used in DETR. Based on track and detectiom queries, TransTrack predicts both detection and track bounding boxes. A simple IoU-based association is used to match the predicted detection and track bounding boxes. In the following frame, associated objects can be used as track queries.

Both detection and track predictions are trained using a loss similar to DETR, which combines the classification loss, bounding box L1 loss, and GIoU (Generalized intersection over union) loss~\citep{giou} (instead of the IoU loss). TransTrack can be trained on static images by simulating the previous frame with random scaling and translations. This allows the transformer-hungry model to be trained on a combination of MOT and object detection datasets (e.g., CrowdHuman~\citep{crowdhuman}).

Compared to traditional tracking-by-detection approaches, it reduces heuristic complexity by eliminating the need for advanced association methods, a re-identification model, or a KF for bounding box prediction. However, unlike fully end-to-end tracking methods, it still requires a separate association step using IoU.

\subsection{Heuristic-based tracking methods}

Heuristic-based methods rely solely on association heuristics to improve association accuracy. They do not introduce model-related improvements and are based on the separate detection-embedding (SDE) architecture.

\textbf{ByteTrack and its derivativs}. Due to its simplicity and effectiveness, \emph{ByteTrack} has become one of the most influential tracking-by-detection methods, serving as the foundation for numerous modern approaches~\citep{imprasso, botsort, bytev2, hybridsort, motiontrack_byte_cmc, deepmovesort, deep_eiou}. Its widespread adoption highlights its role as \emph{the modern SORT}, providing a robust and efficient baseline that researchers continue to refine and extend. The method’s adaptability has led to various enhancements, enabling better handling of challenges such as occlusions, identity switches, and long-term tracking. As a result, ByteTrack and its derivatives are now central to state-of-the-art multi-object tracking. In the following paragraphs, we provide a detailed discussion of ByteTrack’s key innovations, its impact on the field, and the various improvements introduced by its successors.

\textit{ByteTrack}. Recall that SORT's performance is sensitive to the detection score threshold \(det_{\tau}\). Using a high detection threshold reduces false positives but increases false negatives, and vice versa. ByteTrack introduces a simple association method, \emph{Byte}, that allows the use of most detections. The association process is split into two cascaded steps. In the first step, high-score detections (defined by \(det_{\tau}\)) are associated with active and lost tracks based on motion and appearance similarity. In the second step, low-score detections are associated with the remaining (unassociated) active tracks based on motion similarity only. The main motivation is that occluded objects usually have low detection scores, as the model is less confident in these cases. Consequently, in the first step, \emph{clearly visible} objects are prioritized for association, and subsequently, less confident detections are considered. In addition to improving tracker performance~\citep{bytetrack}, the Byte association process is less sensitive to the detection score threshold compared to SORT.  

\begin{figure*}
  \centering
  \includegraphics[width=0.9\linewidth]{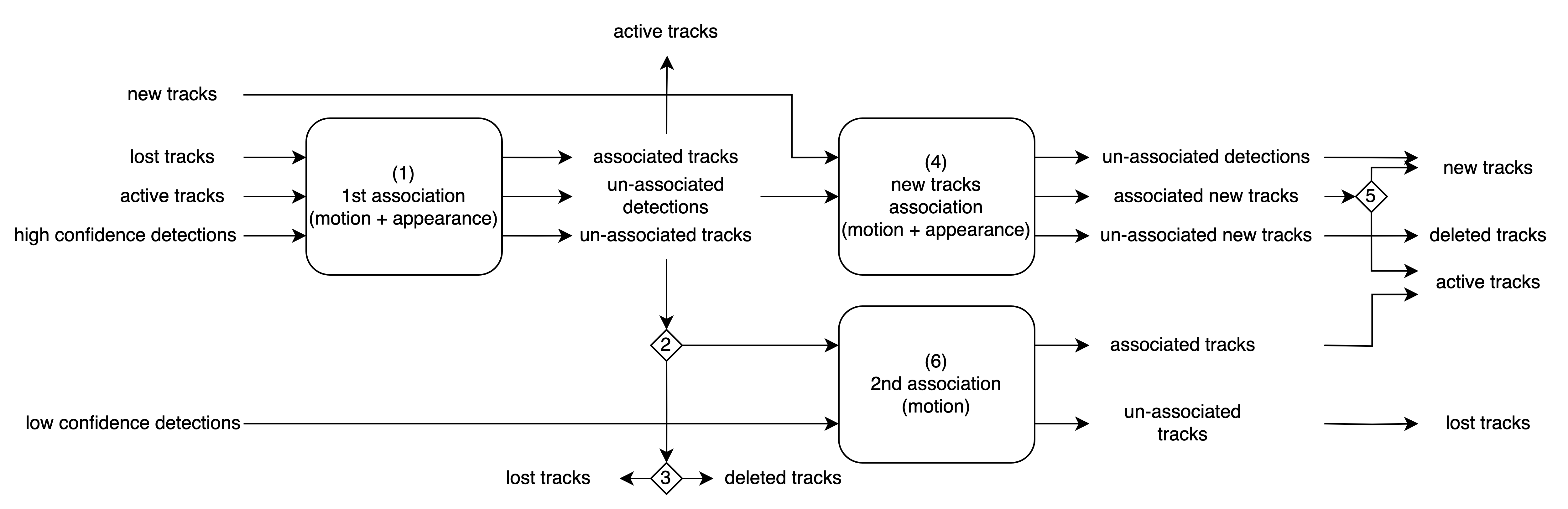}
  \caption{ByteTrack association design. It consists of six steps, marked with numbers 1 to 6. The first association step (1), takes as input tracks with active or lost state and high-score detections. The output of this step includes associated tracks (which are further considered as active tracks), unassociated high-score detections (which are further used in step (4)), and unassociated tracks (which are further used in step (6)). Unassociated tracks are split into active and lost tracks in step (2). Lost tracks are further processed in step (3) to determine whether they should be deleted. Step (4) handles tracks in the probationary period (new tracks). Unassociated detections from this step can be used as potential new tracks. Unassociated tracks are deleted as they failed during the probationary period, while associated tracks are either converted to active tracks or kept as new tracks if their probationary period has not ended (step 5). Step (6) is the second cascade of the \textit{Byte} association approach. Active tracks that were not associated in step (1) are attempted to be associated with low-score detections. Successfully associated tracks become active, while unsuccessfully associated tracks become lost.}
  \label{fig:bytetrack}
\end{figure*}

ByteTrack uses IoU similarity, i.e., motion-based association, in all association cascades. A re-ID model can be used to extend the first association cascade with appearance similarity, but it is not ideal for the second association cascade. This is because low-score detections used in the second association cascade are usually occluded and have poor appearance descriptors~\citep{bytetrack}. Additionally, using a re-ID model in the second cascade would slow down inference, as it would be applied to a large number of bounding boxes. Beyond the high-score and low-score detection association steps, ByteTrack also includes a third association step for tracks in the probationary period, similar to Deep SORT. This association step takes as input new tracks and high-score detections that were not associated in the previous step. A detailed visualization of ByteTrack's association process is shown in Figure~\ref{fig:bytetrack}.  

Apart from introducing Byte, ByteTrack also employs the YOLOX object detector, which has become the default tracker for the MOTChallenge benchmark with private detections~\citep{bytetrack, dancetrack, sportsmot}. Further developments of ByteTrack are introduced in ByteTrackV2~\citep{bytev2}, which unifies ByteTrack for both 2D and 3D MOT settings.  

\textit{BoT-SORT}. BoT-SORT~\citep{botsort} further improves ByteTrack's performance with a few bag-of-tricks. First, BoT-SORT uses Deep SORT's adaptive KF with time-dependent process and measurement noise approximations that scale relative to the object's bounding box width and height. Specifically, KF predictions are more confident for smaller, distant objects that typically exhibit less displacement between frames. Second, BoT-SORT applies a camera motion compensation (CMC) method to correct KF predictions between frames in non-static camera scenes. Conretely, the global motion compensation (GMC) technique is used from OpenCV's Video Stabilization module~\citep{opencv_library}. Additionally, it integrates FastReID's BoT-SBS50~\citep{fastreid_bot}, a stronger re-identification model. This combination of techniques results in improved performance compared to the ByteTrack baseline~\citep{botsort}. Lastly, BoT-SORT changes the Deep SORT's association cost defined in~\eqref{eq:deep_sort_cost} with the new proposed one: 
\begin{equation}
    C_{\text{BoT-SORT}}(i, j) = min\{d_{M}(\boldsymbol{D}_i, \hat{\boldsymbol{T}}_j), d_{A}(\boldsymbol{E}_i, \hat{\boldsymbol{E}}_{j})\}, \label{eq:bot_sort_cost}
\end{equation}
where $d_{A}(\boldsymbol{E}_i, \hat{\boldsymbol{E}}_j)$ is defined as:
\begin{equation}
    d_{A}(\boldsymbol{E}_i, \hat{\boldsymbol{E}}_j) = 
        \begin{cases} 
            0.5\cdot d_{cos}(\boldsymbol{E}_i, \hat{\boldsymbol{E}}_j), (d_{cos}(\boldsymbol{E}_i, \hat{\boldsymbol{E}}_j) < t^{(A)}) \And (d_{M}(\boldsymbol{D}_i, \hat{\boldsymbol{T}}_j) < t^{(P)}), \\
            1, \text{otherwise}
        \end{cases}
\end{equation}
where $d_{\text{cos}}$ is the cosine distance, $\boldsymbol{D}_i$ is $i$th detection bounding box with corresponding appearance feature embdding $\boldsymbol{E}_i$, $\hat{\boldsymbol{T}}_j$ is $j$th track's KF predicted bounding box, $\hat{\boldsymbol{E}}_j$ is the $j$th track's EMA appearance feature embedding, $t^{(A)}$ is the appearance similarity threshold, and $t^{(P)}$ is the proximity threshold. Essentially, the proximity threshold is used to disable appearance-based association in cases where objects are not physically close enough. BoT-SORT also uses EMA for $\hat{\boldsymbol{E}}_j$ estimation~\citep{botsort}.

\textit{ImprAsso}. Although ByteTrack's two-stage association process is highly effective, it has certain limitations.  
First, high-confidence detections are prioritized over low-confidence detections without considering the distances between low-confidence detections and tracks. Consequently, a detection that does not perfectly align with a track's prediction may be preferred over a lower-confidence detection that provides a better match. Second, only active tracks, not lost tracks, are used in the second stage, which can lead to suboptimal association.  ImprAsso~\citep{imprasso} addresses these limitations by introducing a novel \emph{combined matching} approach as an alternative to Byte. This approach integrates both high- and low-confidence detections into a single association step. Since different distance metrics can be applied to high- and low-confidence detections (e.g., appearance similarity may not be used for low-confidence detections), the distances for low-confidence detections are scaled by a factor $\beta = \frac{d^h_{max}}{d^l_{max}}$, where $d^h_{max}$ is the maximum distance value for high-confidence detections with any track, and $d^l_{max}$ is the maximum distance for low-confidence detections with any track.  

In addition to the combined distance association method, ImprAsso proposes an occlusion-aware initialization strategy. New tracks are initialized only if they do not excessively overlap with existing tracks, as defined by an IoU threshold. The authors argue that most false-positive detections in MOT are, in fact, duplicate detections of the same person~\citep{imprasso}.

ImprAsso combines a bag of tricks from previous works. Specifically, it uses Deep SORT's association cost function but replaces the Mahalanobis distance with IoU and employs a gate function based solely on IoU. Additionally, it uses the same re-ID model as BoT-SORT --- FastReID's BoT-SBS50—and the same CMC algorithm.

\textbf{Alternative Cascaded Association Methods}. Association methods generally benefit from cascades to improve tracking accuracy.  
Cascades are typically applied either through a two-stage process~\citep{cbiou, twix, smiletrack, ocsort} or iteratively~\citep{sparsetrack, deep_eiou}. These types of heuristics are not limited to heuristic-based methods, as they are also used in motion-based trackers~\citep{motiontrack_byte_cmc, ocsort, deepmovesort} and affinity learning-based methods~\citep{smiletrack, twix}.

\textit{SparseTrack}. SparseTrack~\citep{sparsetrack} extends the BoT-SORT method by introducing cascaded associations based on objects' pseudo-depth, i.e., the vertical position of bounding boxes, which can be used to estimate depth due to perspective projection. Objects are first sorted based on their pseudo-depth and then split into multiple cascaded levels for association. The association cost between objects is calculated using the standard IoU cost. SparseTrack's heuristic is highly effective in heavily crowded scenes (e.g., MOTChallenge), resulting in high tracking accuracy 
without requiring a re-ID model.

\textit{C-BIoU}. C-BIoU~\citep{cbiou} proposes bounding box shape buffering (expansion) during association, i.e., BIoU, which increases tracking robustness to irregular motions and similar appearances. The expanded bounding box allows association based on the IoU of non-overlapping bounding boxes and compensates for motion model prediction inaccuracies. To reduce the risk of overexpansion, C-BIoU uses cascaded association with two increasing levels of expansion. This heuristic has proven to be very effective on datasets with non-linear motion~\citep{soccernet, sportsmot, dancetrack}. Deep-EIoU generalizes C-BIoU by replacing the fixed two-level cascade with an arbitrary number of levels with incremental expansion.

\textbf{Confidence-Based Association Methods}. Since occluded objects tend to have lower detection confidence compared to non-occluded ones, heuristics based on detection confidence can be employed during the association process. We refer to association methods that utilize such heuristics as confidence-based association methods, as they differ from motion-based and affinity-based methods. ByteTrack was one of the first methods to incorporate detection confidence during association. However, it can be further leveraged to improve association accuracy and reduce identity switches.

\textit{Hybrid-SORT}. Hybrid-SORT~\citep{hybridsort} combines BoT-SORT's baseline and motion-based heuristics from OC\_SORT~\citep{ocsort} with weak cue heuristics for more efficient occlusion handling. Hybrid-SORT introduces \textit{tracklet confidence modeling}, where a KF is used to model bounding box confidence in addition to bounding box coordinates. This can be leveraged to associate tracks and detections based on the detection model's confidence scores. Generally, occluded objects tend to have lower detection confidence, so incorporating confidence in the association step can help reduce the number of ID switches. These heuristics have shown great success on crowded datasets~\citep{mot_challenge}, as well as datasets with non-linear motion~\citep{dancetrack}. Furthermore, Hybrid-SORT proposes \textit{height-modulated IoU (HMIoU)}, which is defined as the standard 2D IoU multiplied by height-based IoU. The motivation behind this heuristic is that objects farther from the camera in a perspective projection tend to appear smaller, even if they have a similar shape in space. The same principle can be applied to width; however, height tends to be a more stable cue~\citep{hybridsort}. 

\textit{BoostTrack}. BoostTrack~\citep{boostrack} introduces a tracking algorithm with a rich set of heuristics designed for crowded scenes. It employs track-detection confidence boosting (a multiplier) for similarity measures, based on the assumption that tracks and detections with high confidence scores tend to have more accurate geometric and appearance features. Track confidence is modeled as an exponential decay, meaning that lost tracks and recently initialized tracks have lower confidence scores. Track-detection confidence is computed as the product of track confidence and detection confidence, which is then used to scale the employed similarity measures based on bounding box overlap and shape differences. Additionally, BoostTrack successfully incorporates the Mahalanobis distance, originally used in Deep SORT, while mitigating its drawbacks related to the uncertainty of lost tracks. This is achieved by applying softmax to the set of Mahalanobis distances between a single track and all detections. Working with normalized distances allows the model to leverage KF's prediction uncertainty without relying on Deep SORT's cascaded association based on a track's lost time. In summary, BoostTrack combines Mahalanobis distance and IoU between bounding boxes, along with scaled and L1 distances between bounding box shapes. Finally, BoostTrack introduces heuristics to increase the confidence of likely objects that may have been occluded and to decrease the confidence of unlikely objects that could be outliers. Specifically, detections with low confidence scores that have high IoU with any existing track have their confidence scores increased. Conversely, unlikely detections are ignored if their minimum Mahalanobis distance to any track exceeds a defined threshold. This combination of heuristics enables BoostTrack to achieve strong performance in extremely crowded scenes, such as those in MOT20~\citep{mot_challenge, boostrack}.

BoostTrack++~\citep{boostrackplus} builds upon BoostTrack with an additional set of refinements, including buffered (expanded) bounding boxes~\citep{cbiou} and further tuning of existing heuristics. These improvements lead to a slight performance gain on the MOTChallenge benchmark compared to the BoostTrack baseline~\citep{mot_challenge, boostrackplus}.

\textbf{Object detection model exploitation}. Tracktor~\citep{tracktor} improves tracking accuracy without requiring additional training by exploiting the Faster R-CNN detector during the association step. Specifically, Tracktor applies the detector's regressor to the last track's bounding box to realign it based on the features from the current frame. This approach is effective in cases where the object has moved only slightly between frames, which is often true for videos with high FPS and slow-moving objects (e.g., pedestrians). The regression precision can be further improved by using the KF's prediction instead of the last bounding box. Additionally, Tracktor uses the detector's classification score to decide whether a track should be terminated; a track is terminated if its classification score falls below a predefined threshold. Finally, Tracktor employs a heuristic to initialize new tracks only when no existing tracks are nearby, specifically, when the IoU between a detection (potential track) and active tracks is below a given threshold. What makes Tracktor interesting is that it requires no additional components or models. Instead, it only smartly exploiting the region-based detector --- Faster R-CNN. 

\subsection{Motion-based tracking methods}

Trackers using the linear constant velocity KF are state-of-the-art solutions on MOTChallenge~\citep{mot_challenge, boostrackplus, botsort, imprasso}, which predominantly features linear motion (aligned with the KF's assumptions). However, linear motion assumption is not valid in the general case. This is precisely why methods based on the KF underperform on datasets with dynamic and non-linear motion~\citep{dancetrack, sportsmot, deepmovesort, ocsort, motiontrack, ettrack}. We refer to methods that address these issues by improving the KF or replacing it with deep-learning-based models as motion-based tracking methods.

\textbf{Observation centric approach}. OC\_SORT~\citep{ocsort} addresses the limitations of the KF through an observation-centric\footnote{In the context of MOT, KF's observations are the output bounding boxes generated by object detection models.} approach. It introduces observation-centric re-update (ORU), which, instead of relying solely on the KF's linear state estimation (i.e., an estimation-centric approach), handles occlusions by re-initializing the KF through interpolation of missing bounding boxes after lost track's re-identification. In other words, if a lost track is successfully associated after $N$ frames, then the track's missing trajectory is approximated through interpolation. After interpolation, the KF's state is updated for the last $N$ frames based on the interpolated trajectory. This process corrects error accumulation in the KF state. The method is motivated by the observation that object detectors perform per-frame detection independently and do not accumulative error over time. In addition to ORU, OC\_SORT proposes two more heuristics: observation-centric momentum (OCM) and observation-centric recovery (OCR). The OCM heuristic augments the association cost function by incorporating motion momentum approximated through observation history as a similarity measure. The motivation is that, over short time intervals, tracks should be associated with detections that would maintain track's consistent motion patterns. The OCR heuristic adds an extra association step for tracks that remain un-associated, matching them with un-associated detections based on their last observed positions rather than the KF's predicted bounding boxes. OC\_SORT serves as a robust alternative to ByteTrack in modern applications, particularly in scenarios where objects exhibit non-linear motion patterns (e.g. DanceTrack or SportsMOT datasets)~\citep{ocsort, dancetrack, sportsmot}.

Deep OC\_SORT~\citep{deepocsort} extends OC\_SORT with CMC (similar to BoT\_SORT), a re-ID model, and \emph{dynamic appearance} features. Instead of using moving average to aggregate track feature embeddings over time, Deep OC\_SORT introduces a modification where momentum is not constant but instead depends on the detection score (dynamic appearance). This adjustment is reasonable, as a low detection score often results from occlusion, leading to poor-quality appearance features. The Deep OC\_SORT's track's appearance update is as follows:
\begin{equation}
    \boldsymbol{E}^{T}_{j, f+1} = (1-\alpha_{f})\cdot \boldsymbol{E}^{D}_{j, f+1} + \alpha_{f} \cdot \boldsymbol{E}^{T}_{j, f} \label{eq:deepocsort_appearance_cost}
\end{equation}
where $\alpha_{f}$ is the moving average momentum for $f$th frame computed as:
\begin{equation}
    a_{f+1} = a_{f} + (1-a_{f})\cdot(1-\frac{c_{i}-det_{\tau}}{1-det_{\tau}}) \label{eq:deepocsort_alpha_update}
\end{equation}
where $c_i$ is the confidence score of the $i$th detection, and $det_{\tau}$ is the detection score threshold used to filter detections. This modification does not introduce any new hyperparameters while making the appearance update more robust to low-quality appearance embeddings. In summary, Deep OC\_SORT extends OC\_SORT by incorporating missing components required for state-of-the-art performance on popular benchmarks.

\textbf{Deep learning-based motion models}. MotionTrack~\citep{motiontrack_byte_cmc} employs a transformer-based motion model that considers an object's history as well as its interactions with other objects. The motion model consists of two modules: an interaction module for short-range association and a refind module for long-range association. Short-range association is used for active tracks, while long-range association is used for lost tracks. The interaction module predicts an object's motion based on its historical geometric, motion, and interaction features related to other objects present on the scene. Interaction features are extracted using a combination of the attention mechanism~\citep{attention_is_all_you_need} and asymmetric convolutions~\citep{acnet}, while their fusion is performed with a graph convolutional network~\citep{gcn}. For the refind module, asymmetric convolutions are also used to extract features which are then used to construct a correlation matrix between detections and lost tracks. This correlation matrix serves as input to a linear assignment process to identify tracks that are \textit{refound}. Replacing the KF with a deep learning-based motion model that considers object interactions led to strong performance on the MOTChallenge dataset at the time. 

A tracker with the same name~\citep{motiontrack} and a similar nature introduces a transformer-based motion predictor as an alternative to linear models often used in MOT. This motion predicted shows great performance when trained on a large-scale dataset --- SportsMOT~\citep{sportsmot}. This tracker performs very well on the DanceTrack~\citep{dancetrack} and SportsMOT~\citep{sportsmot} datasets compared to methods that use neither a learnable motion predictor nor a strong re-ID model. However, achieving this performance requires additional datasets~\citep{sportsmot}.

ETTrack~\citep{ettrack} is another tracker based on a learnable motion predictor that combines transformers~\citep{attention_is_all_you_need} and temporal convolutional network (TCN) blocks~\citep{tcn}. This method emphasizes motion momentum by introducing a momentum correction loss, a regularizer that tasks the model with predicting an object's direction as well. Combined with OC\_SORT's momentum association heuristic, ETTrack demonstrates strong performance on datasets with non-linear motion~\citep{dancetrack, sportsmot}.

Another concurrent work based on a learnable motion model is MoveSORT~\citep{movesort}. MoveSORT replaces the commonly used KF with a deep learning-based filter that offers the same set of features, such as uncertainty estimation and noise filtering. Two families of deep learning-based filters are proposed: Bayesian and end-to-end. Bayesian filters can be used when the measurement likelihood is known or can be accurately estimated; otherwise, end-to-end variants can be used, as they simultaneously learn to predict an object's motion and filter detection noise originating from an inaccurate detection model.  Additionally, MoveSORT introduces a hybrid association cost that combines IoU similarity and L1 distance between bounding boxes during association. Essentially, L1 distance accounts for both bounding box scale and absolute position. Similar to MotionTrack~\citep{motiontrack} and ETTrack~\citep{ettrack}, MoveSORT performs well on the DanceTrack and SportsMOT~\citep{sportsmot} datasets, as it was designed for such cases. 

DeepMoveSORT~\citep{deepmovesort} extends MoveSORT with an improved transformer-based end-to-end filter, a re-ID model, and additional heuristics, achieving state-of-the-art performance on datasets with non-linear motion~\citep{dancetrack, sportsmot}. However, simplicity and elegance are sacrificed compared to the MoveSORT tracker due to extensive engineering aimed at achieving strong results. This emphasizes the drawback of the standard tracking-by-detection paradigm, as many tuned components and hyperparameters are required for strong performance.

\textbf{Ground-plane motion model}. UCMCTrack~\citep{ucmctrack} addresses challenges in video sequences with significant camera movements, in which case CMC and a re-ID model are crucial~\citep{botsort, sparsetrack}. The authors propose a motion-based tracker that is robust to camera movements without requiring CMC or a re-ID model. Unlike traditional methods that operate in the image plane, UCMCTrack projects objects onto the ground plane as points. Objects are projected based on their bottom-center bounding box coordinate. They use a KF with a constant velocity assumption for the ground-plane motion prediction, analogous to methods working in the image plane. They approximate measurement noise as proportional to bounding box width and height, as in~\citep{deep_sort, botsort}. For process noise, they integrate camera movement. For association, they use the normalized Mahalanobis distance mapped to the ground plane, which increases distance based on uncertainty magnitude:
\begin{equation}
    \boldsymbol{D} = \boldsymbol{e}^T \boldsymbol{S}^{-1} \boldsymbol{e} + \ln |\boldsymbol{S}| \label{eq:ucmctrack_mahalonobis}
\end{equation}
where $\boldsymbol{e}$ is the residual between the KF's prediction and a detection (measurement), $\boldsymbol{S}$ is the corresponding covariance matrix (uncertainty), and $|\boldsymbol{S}|$ represents the determinant of $\boldsymbol{S}$. The first summation component is equal to the squared Mahalanobis distance, while $\ln |\boldsymbol{S}|$ increases the distance based on uncertainty. This element is crucial, as it partly resolves Deep SORT's issue related to the use of the Mahalanobis distance and increased uncertainty (which otherwise decreases distance) for lost tracks.

UCMCTrack demonstrates state-of-the-art performance on multiple datasets~\citep{kitti, mot_challenge, dancetrack} with a unique approach and without using any re-ID model. However, there are some practical limitations to consider. First, while the method aims to avoid CMC, its benefits remain significant on MOT17~\citep{ucmctrack}, where camera movement introduces considerable noise compared to other datasets. This suggests that in some scenarios, CMC may still be beneficial despite the method's intended design. Second, the approach requires manually tuned camera parameters for each scene~\citep{ucmctrack}, which limits its adaptability to unseen environments. In other words, manual parameter tuning is necessary at test time, unlike many other methods that do not require such adjustments.

\subsection{Affinity learning-based tracking methods}

Affinity learning-based tracking methods improve data association by learning expressive similarity functions between detections and tracks. Unlike traditional approaches relying on handcrafted metrics, they adapt to appearance variations, occlusions, and identity switches without requiring domain-specific motion assumptions. By leveraging contrastive learning, metric learning, or graph-based models, these methods enhance tracking robustness.

\textbf{Motion affinity learning}. The online variant of StrongSORT~\citep{strongsort} combines a multiple heuristics to improve the Deep SORT baseline. This includes using an improved appearance model like in ~\citep{botsort}, moving average for track appearance history aggregation like in~\citep{jde}, a NSA KF~\citep{giaotracker} for motion prediction, and ECC (enhanced correlation coefficient maximization)~\citep{ecc} for CMC. However, the offline variant—StrongSORT++—is what makes this tracking method particularly interesting. An AFLink module is trained and used to globally link tracks, addressing missed associations. This model relies only on a track's coordinate trajectory. Trajectory features are extracted using multiple layers of 1D CNNs. The affinity between two tracks is computed by concatenating their embeddings and using an MLP to predict whether they should be associated, i.e., their connectivity. AFLink is highly efficient in resolving missed associations; however, it cannot mitigate ID switches~\citep{strongsort}. In addition to AFLink, StrongSORT++ performs interpolation using Gaussian Process Regression (GPR)~\citep{gpr} instead of standard linear interpolation. Compared to linear interpolation, GPR can more accurately interpolate tracks with non-linear motion.

TWIX~\citep{twix} introduces an affinity module trained to detect pairs of tracks belonging to the same object. This is achieved through a contrastive learning framework using only a track's spatio-temporal coordinates, i.e., without appearance features. First, pairs of tracklets are encoded using a transformer with attention applied over the temporal dimension (intra-pair encoding). Second, another transformer enriches features based on other track pairs (inter-pair encoding). Finally, a linear layer predicts the pair affinity score. During inference, TWIX calculates affinity scores between historical tracks and detections. More precisely, an extended version, C-TWIX, is used, which employs cascaded matching similar to C-BIoU~\citep{cbiou}, using two different TWIX modules in the first and second association steps. What makes TWIX particularly notable is that it does not require IoU for motion-based association or any domain-specific heuristics, yet it maintains performance comparable to state-of-the-art methods across multiple datasets~\citep{twix}.

\textbf{Appearance affinity learning}. SMILETrack introduces a similarity learning module (SLM) that extracts object appearance features and predicts appearance similarity between different objects. Specifically, detection bounding boxes are used to crop object features from the frame, which are then processed using a vision transformer-based network~\citep{vit}. The SLM module can be used instead of the standard re-ID model with cosine distance for appearance-based association. Apart from appearance-based association, SMILETrack's tracking algorithm is similar to ByteTrack. Compared to the previous state-of-the-art at the time (BoT-SORT), SMILETrack improved tracker association accuracy~\citep{smiletrack}.

QDTrack~\citep{qdtrack} introduces a quasi-dense similarity learning approach for multi-object tracking. Unlike traditional methods that rely on sparse ground truth matching, QDTrack densely samples numerous object regions from pairs of images to perform contrastive learning. This strategy focuses solely on appearance features, eliminating the need for motion prediction. In this framework, each detection is processed to extract appearance embeddings, which are then compared across frames using a nearest neighbor search to associate objects over time. The quasi-dense sampling enables the model to learn a more discriminative feature space, enhancing its ability to distinguish between different objects based on appearance alone. Notably, QDTrack's similarity learning scheme is not limited to video data; it can effectively learn instance similarity from static images, allowing for competitive tracking performance without requiring video-based training or tracking supervision. Experimental results demonstrate that QDTrack achieves performance comparable to state-of-the-art methods across multiple benchmarks, including setting a new standard on the BDD100K~\citep{bdd100k} MOT benchmark, all while introducing minimal computational overhead to the detector~\citep{qdtrack}.

\subsection{Offline tracking methods}

Offline tracking-by-detection methods require all detections to be generated beforehand, performing data association globally rather than frame-by-frame. This allows for optimal assignment strategies but prevents real-time processing. As a subset of affinity learning-based tracking, these methods construct graphs where detections are nodes and relationships are learned through similarity functions, often using Graph Neural Networks (GNNs)~\citep{gnn_survey}. The GNN methods show incredible performance~\citep{sushi, cono_link} but can only be used for offline tracking. However, it's important to note that almost every ``online'' method uses linear~\citep{bytetrack, botsort, sparsetrack, smiletrack, imprasso, boostrack, boostrackplus, movesort, deepmovesort, cbiou, hybridsort, twix, ettrack} or GSI (Gaussian Smoothed Interpolation)~\citep{strongsort, ocsort, deepocsort} and hence the results presented are almost always with offline post-processing.

\textit{MPNTrack}. MPNTrack~\citep{mpntrack} formulates MOT as a graph-based assignment task, where link prediction is used to associate detections across frames. In this graph, detections serve as nodes, and edges represent connections between them. MPNTrack employs a GNN with a message-passing network (MPN) architecture to model pairwise relationships between detections. The model extracts appearance features from image patches corresponding to detection bounding boxes using a CNN, which are then used as initial node features. Edge features are defined as a combination of relative geometric coordinates, appearance similarity, and time differences between detections. These node and edge features serve as inputs to the GNN framework. A message-passing network is applied to iteratively exchange information between nodes, where each step enriches nodes and edges with information from their neighbors. 

The node-to-edge information update process is defined as:
\begin{equation}
    \boldsymbol{h}^{(l)}_{i, j} = f_{e} (\boldsymbol{h}^{(l-1)}_{i}\;||\;\boldsymbol{h}^{(l-1)}_{j}\;||\;\boldsymbol{h}^{(l-1)}_{i, j}) \label{eq:mpntrack_node_to_edge}
\end{equation}
where $l$ is the GNN layer corresponding to the step number, $i$ is the index of the $i$th detection with the corresponding node features $\boldsymbol{h}^{(l-1)}_{i}$, $(i, j)$ is the edge between the $i$th and $j$th detections with corresponding edge features $\boldsymbol{h}^{(l-1)}_{i, j}$, and $f_{e}$ is a learnable function, specifically a multi-layer perceptron (MLP). The edge-to-node update process is defined as:
\begin{equation}
    \boldsymbol{m}^{(l)}_{i, j} = f_{v} (\boldsymbol{h}^{(l-1)}_{i}\;||\;\boldsymbol{h}^{(l)}_{i, j}) \label{eq:mpntrack_edge_to_node}
\end{equation}
where $f_{v}$ is another learnable MLP function. The final node embeddings are obtained through an aggregation step:
\begin{equation}
    \boldsymbol{h}^{(l)}_{i} = \Phi(\{\boldsymbol{m}^{(l)}_{i, j}\}_{j\in \boldsymbol{N}_{i}}) \label{eq:mpntrack_aggregation}
\end{equation}
where $\boldsymbol{N}_{i}$ represents the neighbors of the $i$th detection (node), and $\Phi$ is an order-invariant aggregation function, such as summation. Equations~\eqref{eq:mpntrack_node_to_edge},~\eqref{eq:mpntrack_edge_to_node}, and~\eqref{eq:mpntrack_aggregation} define the message-passing framework. To account for temporal dependencies, MPNTrack aggregates information separately from past and future detections in~\eqref{eq:mpntrack_edge_to_node} using different networks before combining them in a \textit{time-aware message-passing} mechanism.

Since the graph scales exponentially with video length, an edge between two nodes is retained only if both are among the top $50$ mutual nearest neighbors based on appearance similarity. MPNTrack’s joint modeling of visual and geometric cues improves robustness to occlusions and reduces identity switches. The method processes video clips of $15$ frames with a stride of $1$ and merges predictions post-processing~\citep{mpntrack}.

\textit{SUSHI}. SUSHI~\citep{sushi} extends MPNTrack by replacing the monolithic graph structure with a hierarchy of graphs. In the first step, pairs of frames are associated based on their relative detection features. A SUSHI block is used to associate detections, resulting in tracks with a maximum length of $2$. In the next step, a graph of these tracks is formed, with the maximum length of each track extended to $4$. This process is repeated $9$ times, producing tracks with a maximum length of $512$ frames. The SUSHI block is essentially similar to MPNTrack, except it additionally incorporates motion features as inputs alongside geometric and appearance features. By sharing weights across the SUSHI blocks, the number of training samples is effectively increased, leading to empirically improved performance and better convergence. Similar to MPNTrack, only the $10$ nearest neighbors are considered for each node based on geometric, motion, and appearance similarity. Videos are processed offline as clips of length $512$ with a stride of $256$. These modifications to the original MPNTrack method result in a significant performance boost across multiple datasets~\citep{sushi}.

\textit{CoNo-Link}. When formulating MOT as a graph-based assignment problem, the size of the graph grows exponentially with video length. Learning associations through these graphs is infeasible in terms of both computation and memory~\citep{mpntrack, sushi, cono_link}. To mitigate this, edge pruning based on appearance features is necessary. In the case of MPNTrack, retaining the top 50 neighbors per node is required to maintain 99.9\% envelopment. The risk of mistakenly pruning crucial edges still exists even if many neighbors are retained. Instead, CoNo-Link introduces \textit{NodeNet}, a transformer-based neural network that models the association likelihood between two detections (or a track and a detection). Using these likelihoods, a pruned graph is formed, preserving only essential edges (e.g. $5$ to $10$). This pruned graph is constructed over a video clip using a tracking-by-detection approach, where tracks from previous frames (with their detections) are associated with detections in the current frame. The similarity between a track and a detection is computed as the sum of all association scores between the track’s detections and the detection in the current frame. In addition to the NodeNet association score, IoU similarity is also incorporated. The resulting graph is used as input to a GNN, which integrates tracks obtained from NodeNet into full video tracks. Compared to other approaches, CoNo-Link incorporates detection-to-detection, detection-to-track, and track-to-track edges in the constructed graph. The final tracking solution is derived by aggregating track-to-track links. CoNo-Link is a more complex solution compared to MPNTrack and SUSHI, but it outperforms them on multiple datasets~\citep{cono_link}.

\subsection{Tracking-by-detection discussion}

The main benefit of the tracking-by-detection paradigm is its modularity. Each component (e.g., object detector, motion model, association, etc.) can be replaced independently, making it highly adaptable to specific task requirements. This flexibility has resulted in a wide range of tracking methods. For benchmark datasets, a slower and larger object detector, such as YOLOX-X, with many parameters and high resolution, is typically used to achieve the highest tracking accuracy. On the other hand, for tasks requiring real-time performance, a smaller object detection model (e.g., YOLOX-S) can be employed~\citep{yolox}. If objects' motion is noisy due to unpredictable camera movement, a camera motion compensation (CMC) module should be employed~\citep{botsort, deep_sort, strongsort}. In domains where objects exhibit dynamic and nonlinear motion, deep learning-based motion models combined with a strong re-ID model are crucial~\citep{deepocsort, deepmovesort}. Last but not least, association accuracy can be improved with a set of heuristics based on domain knowledge.

The main drawback of current tracking-by-detection methods are their dataset-specific heuristics. For instance, the standard KF is widely used as the default motion model~\citep{sort, deep_sort, bytetrack, botsort, bytev2, ocsort, boostrack, hybridsort, sparsetrack, tracktor}. However, it struggles with modeling non-linear motion and requires domain knowledge for parameter tuning. Additionally, many tracking methods rely heavily on domain-specific association heuristics rather than leveraging general machine learning models capable of learning from the data. 

We argue that this limitation stems from the historically limited variety of datasets available for tracker evaluation. MOTChallenge, the most popular MOT benchmark, is specialized for pedestrian tracking~\citep{mot_challenge}. However, with the introduction of additional large-scale datasets, research is shifting towards more general solutions --- end-to-end tracking~\citep{motr, motrv2, motrv3, memot, memotr, motip}.

We argue that affinity learning, including offline GNN-based methods, represents the most natural progression for the tracking-by-detection paradigm from the perspective of deep learning and machine learning. These methods are capable of learning directly from data while preserving modularity and flexibility through a separate object detection model.

\section{End-to-end tracking paradigm}

We define a tracking method as end-to-end if object detection and association are trained jointly. Such methods require no post-processing after association, and the only tracker logic and hyperparameters pertain to track handling (track initialization, track deletion, etc.). Currently, most well known end-to-end multi-object tracking methods are extensions of the DETR~\citep{detr} end-to-end object detection model~\citep{motr, motrv2, motrv3, memot, memotr, motip, trackformer}, which was detailed in Section~\ref{sec:tbd}. We first introduce MOTR (end-to-end Multi-Object Tracking with Transformer)~\citep{motr} along with its modifications to DETR. This provides a foundation for understanding other end-to-end methods, which are similar to MOTR. MOTR and related approaches~\citep{motrv2, motrv3, memot, memotr} are commonly referred to as tracking-by-query MOT trackers.

\subsection{End-to-end tracking methods}

\begin{figure*}
  \centering
  \includegraphics[width=0.9\linewidth]{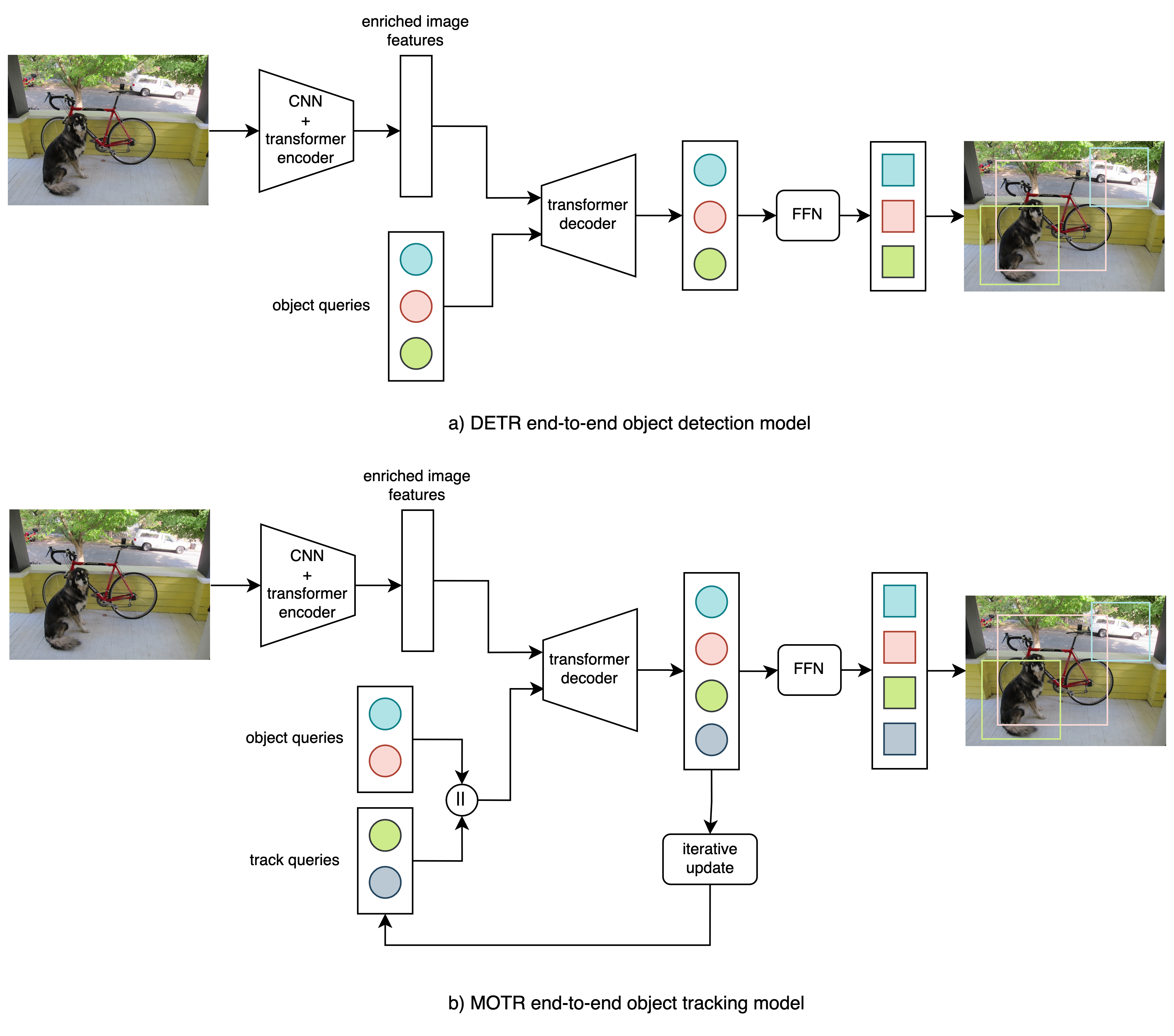}
  \caption{Comparison of DETR and MOTR architectures. (a) DETR uses object queries with a Transformer decoder for end-to-end object detection. (b) MOTR extends DETR by adding track queries, enabling iterative updates for multi-object tracking.}
  \label{fig:detr_motr}
\end{figure*}

\textbf{MOTR}. We introduce MOTR as an extension of DETR for multi-object tracking. As illustrated in Figure~\ref{fig:detr_motr}, the DETR pipeline proceeds as follows: it takes an image as input and extracts features using a CNN backbone. These features are further enriched using a transformer encoder. DETR's decoder predicts object embeddings via an attention mechanism, using the image features (key and value) and learnable detect queries as input. The predicted embeddings are then used to derive objects' bounding boxes and classes.

As shown in Figure~\ref{fig:detr_motr}, MOTR introduces two key modifications to the DETR architecture: \textit{track queries} and their \textit{iterative updates}. In addition to detect queries, which identify new objects, MOTR extends the query set with track queries that encode an object's historical context. These track queries are responsible for detecting tracked objects in the current frame. At a high level, MOTR can be seen as an extension of DETR that incorporates temporal context from previous frames. As an object's state evolves over time, its corresponding track query is iteratively updated, maintaining a continuously refined hidden state.

Successful detection of a tracked object requires updating its query based on the current frame's features. This is accomplished through a \textit{Temporal Aggregation Network} (TAN), which enhances temporal relationship modeling. Specifically, track queries from the current and previous frames are aggregated using an attention mechanism, followed by an MLP and normalization layers. In theory, this allows the model to aggregate a track's context over time in an autoregressive manner~\citep{motr}.

\textit{Training}. MOTR extends DETR's bipartite matching training approach. For newborn objects, bipartite matching is performed in the same way as in DETR, where detect queries are assigned to the ground truths of newly detected objects. On the other hand, track query assignments are consistent with ground truth IDs from the previous frames. 

Training is conducted on video clips of varying lengths. Predictions from the full clip are used to compute frame losses, which are then averaged before backpropagation. The loss consists of three components, similar to DETR~\citep{detr}: classification, L1, and GIoU loss~\citep{giou}. The final loss is calculated as a weighted sum of these components.

\textit{Inference}. During inference, MOTR processes videos in a fully end-to-end manner, without requiring any additional logic after each frame's inference~\citep{motr}. In the initial frame, only detect queries are used. High-confidence detections are used to initialize tracks and their corresponding track queries. In subsequent frames, the model uses track queries to predict the new positions of tracked objects and detect queries to identify new objects. The model is designed to handle duplicate detections robustly (e.g., the same object should not be detected by both detect and track queries)~\citep{detr, motr}. Additionally, if a track's confidence drops below a certain threshold, it is removed from the track set~\citep{motr}. The number of detection queries is fixed and set to a value larger than the maximum number of objects expected in any given frame of the dataset, while the number of track queries varies dynamically over time~\citep{detr, motr}.

\textbf{TrackFormer}. A method similar to MOTR is TrackFormer~\citep{trackformer}, which relies on NMS post-processing, preventing it from being a fully compact end-to-end solution. Additionally, TrackFormer does not use an advanced iterative update network like TAN and instead relies on embedding predictions from the previous frame. 

TransTrack~\citep{transtrack}, mentioned in Section~\ref{sec:tbd}, shares similarities with TrackFormer and MOTR. The transition from TransTrack to TrackFormer involves merging the decoders, resulting in integrated association, as seen in TrackFormer. The transition from TrackFormer to MOTR entails removing the post-processing step after model inference.

\textbf{MOTRv2}. Compared to tracking-by-detection methods, which rely on standalone object detections, the main limitation of MOTR is its poor detection performance. This issue stems from having too few supervision labels for detection queries, as only newborn objects are 
considered~\citep{motr, motrv2, motrv3}. MOTRv2~\citep{motrv2} addresses this limitation by introducing YOLOX, which generates a set of bounding 
box proposals that can be used as anchors for detection queries within the MOTR framework. In this approach, it is sufficient to predict anchor-relative offsets 
instead of absolute bounding box coordinates. This modification significantly improves MOTR's performance across multiple datasets. However, it represents a step backward from the end-to-end approach, as it requires an independently trained object detection model during inference. MOTRv2 does not satisfy our definition of an end-to-end tracker and cannot be considered one. However, it is included in this section as it is based on MOTR.

\textbf{MOTRv3}. MOTRv3~\citep{motr} overcomes MOTR's limitations without including an additional object detection network during inference, unlike MOTRv2. MOTRv3 is a direct successor to the original MOTR and is independent of the modifications introduced in MOTRv2. The modifications compared to MOTR are as follows.
First, they modify the bipartite matching between (detection and track) queries and ground truths during training. They eliminate the use of track query assignments based on ground-truth IDs from previous frames, relying instead on matching costs such as class labels 
and bounding box distances. This adjustment allows the training process to automatically balance assignments between track and detection queries. As a result, 
detection query assignments are more frequent at the start of training and decrease as training progresses. This leads to a significant 
improvement in detection performance compared to the original MOTR~\citep{motrv3}. Notably, the supervision for the last MOTR decoder layer remains unchanged from 
the original paper. Secondly, MOTRv3 proposes leveraging YOLOX outputs for pseudo-label distillation. Unlike MOTRv2, the additional object 
detection model is only used during training and not during inference. Lastly, inspired by Group DETR~\citep{group_detr}, MOTRv3 employs track 
group queries instead of track queries to further enhance supervision. With these modifications, MOTRv3 achieves tracking performance on par with MOTRv2 while maintaining an elegant end-to-end training and inference framework.

\textbf{MeMOT}. Compared to MOTR, which iteratively updates track queries, MeMOT (Multi-Object Tracking with Memory) ~\citep{memot} uses a spatio-temporal memory buffer that stores the history of track states, i.e., track embeddings. Memory buffer is aggregated using an transformer-based memory aggregator which encodes separately both short-term and long-term temporal dependencies. These dependencies are then aggregated into track queries and memory updates. Detect queries are generated through a transformer decoder proposal network. Finally, detect and track queries, with image features, are used as inputs to the (memory) decoder that predicts track's bounding boxes, objectness, i.e. object visibility, and uniqueness, i.e. if the object is unique. Objectness and uniqueness scores are used in track handling logic to discard background and duplicate objects. 

\textbf{MeMOTR}. MeMOTR~\citep{memotr} is an end-to-end tracker most similar to the MeMOT tracking method. The primary difference is that, instead of a memory buffer, MeMOTR uses a long-term memory embedding --- a running average of the decoder outputs. Unlike MeMOT, where the aggregated memory buffer is used for queries, this long-term memory embedding is used as the key in the transformer decoder, replacing image features.

\textbf{MOTIP}. MOTIP (Multi-Object Tracking as ID Prediction)~\citep{motip} differs from tracking-by-query methods, i.e., MOTR and MeMOT variants, as it addresses the MOT problem from a different perspective. Tracking-by-query methods can be viewed as extended DETR object detectors with context from previous frames. In contrast, given a set of tracks' state histories (DETR output history), MOTIP assigns the ID of the most similar track to each detected object. The MOTIP decoder uses tracks' state histories as keys and values, and DETR outputs, i.e., detected object embeddings, as queries. The decoder classifies the ID of each detection based on the list of all track IDs. MOTIP employs a learnable ID dictionary with keys assigned to each track. Predicting learnable keys, instead of fixed one-hot label representations, enhances the model's usability for long videos, where the total number of IDs can be very large. Assignments based on MOTIP outputs are performed using either greedy assignment or the Hungarian algorithm. We can observe that MOTIP shares its similarity with affinity learning methods. MOTIP's performance is on par with the best tracking-by-query methods~\citep{motip}.

\subsection{End-to-end tracking discussion} 

The mentioned end-to-end tracking methods, primarily related to the tracking-by-query paradigm, offer fully differentiable models that can be jointly optimized for detection, association, and motion modeling in a single training phase. Since these solutions learn to track directly from data, they are not domain-specific and provide a generalizable approach. However, they come with significant computational demands. First, they require large numbers of GPUs with high VRAM~\citep{motr, motrv2, motrv3, memot, memotr, motip} for training, making them inaccessible to most researchers and industry practitioners. Second, they are slow during inference, which prevents their use in real-time applications. Currently, tracking-by-detection methods are generally more suitable for real-time object tracking. Lastly, they generally perform poorly compared to tracking-by-detection methods on heavily crowded scenes~\citep{motr, memot, motrv2, motrv3, memotr, motip}.

\section{Methods comparison}

We compare state-of-the-art methods on the previously mentioned benchmark datasets, dividing them into two groups: (1) highly crowded scenes with linear object motion (MOT17 and MOT20) and (2) moderately crowded scenes with dynamic motion and similar object appearances (DanceTrack, SportsMOT). We categorize MOT methods into three paradigms: \textit{end-to-end}, \textit{tracking-by-detection}, and \textit{offline} (a subset of tracking-by-detection). In this context, \textit{offline} refers to methods that perform association after all detections have been obtained.

\begin{figure*}
  \centering
  \includegraphics[width=1.0\linewidth]{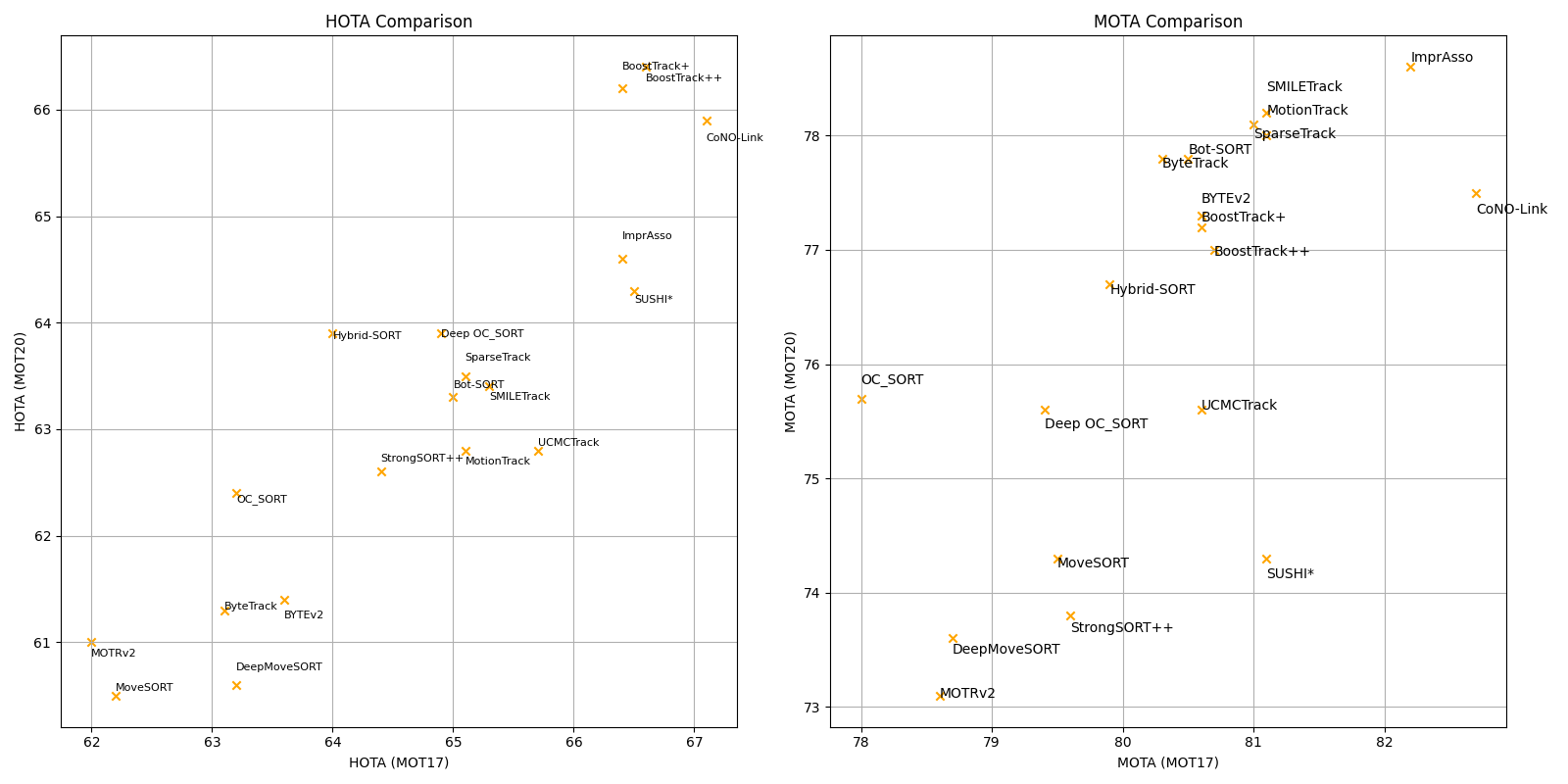}
  \caption{Comparison of HOTA and MOTA scores between MOT17 and MOT20 benchmarks for various tracking methods.
The left plot shows the relationship between HOTA scores, while the right plot highlights MOTA scores. Methods are annotated with their names, and slight vertical offsets help distinguish overlapping points.}
  \label{fig:motchallenge_comparison}
\end{figure*}

\textbf{MOTChallenge benchmark}. Comparison of state-of-the-art methods on MOT17 and MOT20 are presented in Table~\ref{tab:mot17_benchmark} and Table~\ref{tab:mot20_benchmark}, respectively. Additionally, Figure~\ref{fig:motchallenge_comparison} provides a visual comparison of these methods on both MOT17 and MOT20 datasets. Over the past three years (2022--2024), tracking performance in terms of HOTA has improved by $4.0\%$, i.e. from $63.1\%$ (ByteTrack) to $67.1\%$ (CoNo-Link) on the MOT17 dataset and $5.1\%$, i.e. from $61.3\%$ (ByteTrack) to $66.4\%$ (BoostTrack++) on the MOT20 dataset. When considering only online methods, the most improvement has been achieved with the BoostTrack++ tracker, with $3.4\%$ and $5.1\%$ improvements in terms of HOTA on MOT17 and MOT20, respectively. 

From Figure~\ref{fig:motchallenge_comparison}, we observe that BoostTrack++, CoNo-Link, and ImprAsso stand out as they are Pareto-optimal in terms of either HOTA or MOTA. Both BoostTrack++ and ImprAsso are SORT-like methods that achieve strong performance through combination motion-, appearance-, and detection confidence-based associations. These association components are employed through a rich set of heuristics, which allow them to efficiently handle crowded scenes in the MOTChallenge datasets. However, these trackers lack deep learning-based improvements relative to earlier approaches (e.g., ByteTrack, BoT-SORT), as they rely on the same object detection and Re-ID models. The CoNo-Link offline tracking method stands out due to its strong performance, achieved by association techniques combining transformers~\citep{attention_is_all_you_need} and graph neural networks (GNNs)~\citep{gnn_survey}. Unlike BoostTrack++ and ImprAsso, whose performance depends on manually tuned hyperparameters, CoNo-Link's performance scales with dataset size and quality. A noteworthy offline alternative to CoNo-Link is SUSHI, a simpler yet effective approach. 

Based on the results in Table~\ref{tab:mot17_benchmark} and Table~\ref{tab:mot20_benchmark}, we can conclude that end-to-end MOT methods perform worse compared to tracking-by-detection methods on the MOTChallenge benchmark, which includes highly crowded scenes. Results for end-to-end methods are typically reported only on MOT17 and are often missing for MOT20~\citep{motr, motrv2, motrv3, memot, memotr, motip}, which features even more crowded scenarios. However, based on the presented results in Table~\ref{tab:mot20_benchmark} and the similar nature of all end-to-end methods, we can assume that these methods generally perform poorly on MOT20 compared to tracking-by-detection approaches employing YOLOX object detector. The authors of MOTRv2 highlight that one of the main challenges of MOT17 for training transformer-based models is the small dataset size in terms of video duration~\citep{motrv2} — 463 seconds for MOT17 and 535 seconds for MOT20. Additionally, in MOTIP, it is noted that this issue may be related to the limitations of transformer-based detectors when detecting small and densely packed objects~\citep{detr, deformable_detr, motip}. For MOTRv2, the best-performing end-to-end tracker on MOT17, the use of the YOLO object detector for detection proposals partly mitigates DETR's limitations.

Finally, we discuss the effectiveness of the most important association components of tracking-by-detection methods: motion model, re-ID model, CMC (Camera Motion Compensation), and association heuristics. Online state-of-the-art trackers on MOT17 and MOT20 use the standard KF, which includes a linear motion model with a constant velocity assumption~\citep{boostrack, boostrackplus, imprasso, ucmctrack, hybridsort, cbiou, sparsetrack, deepocsort, ocsort, bytetrack, botsort}. As pedestrian motion in this case is linear and predictable, a learnable motion model~\citep{motiontrack, ettrack, movesort, deepmovesort} is expected to be unhelpful for this domain. Replacing the KF with a deep learning-based motion model in this context leads to suboptimal tracker performance, as evidenced by the performance of MoveSORT, ETTrack, and DeepMoveSORT, as well as the ablation studies in~\citep{deepmovesort}. On the other hand, every performant method uses a re-ID model. However, the re-ID model does not have a significant effect in this case, as can be seen when comparing the metrics of BoT-SORT and BoT-SORT-REID in~\citep{botsort}. Using a re-ID model results in $0.4\%$ and $0.7\%$ performance improvement in terms of HOTA on MOT17 and MOT20, respectively. Similar conclusions can be drawn from other ablation studies~\citep{bytetrack, deepocsort, sparsetrack}. A more crucial component, particularly on MOT17 with frequent and unpredictable camera movements, is the CMC~\citep{botsort, deepocsort, deepmovesort, motiontrack_byte_cmc, sparsetrack, boostrack}. Integrating CMC (e.g., GMC from the OpenCV library~\citep{opencv_library}) into a tracker on MOT17 can lead to a $1.0\%$ to $1.5\%$ improvement in terms of HOTA~\citep{botsort, deepocsort, deepmovesort}. Since MOT20 primarily consists of static cameras, CMC is not as relevant~\citep{deepocsort}. Lastly, the addition of a rich set of heuristics significantly enhances tracking performance on both datasets compared to the simple ByteTrack baseline. Some of these heuristics are based on pseudo-depth~\citep{sparsetrack, deepmovesort}, detector confidence~\citep{bytetrack, hybridsort, boostrack, boostrackplus, deepmovesort}, object size~\citep{hybridsort, boostrack, movesort, deepmovesort}, bounding box expansions~\citep{cbiou, deepmovesort}, motion momentum~\citep{hybridsort, ocsort, deepocsort}, and more.

\begin{table*}
\small
\centering
\begin{tabular}{lcccccccc}
Method & Year & HOTA & AssA & DetA & MOTA & IDF1 & IDSW \\
\hline
\textit{offline} & \multicolumn{7}{c}{} \\
SUSHI$^*$~\citep{sushi} & 2022 & \textit{66.5} & \underline{\textit{67.8}} & \textit{65.5} & 81.1 & \textit{83.1} & 1149 \\
CoNo-Link$^{*}$~\citep{cono_link} & 2024 & \underline{\textbf{67.1}}& \underline{\textit{67.8}} & \underline{\textbf{66.7}} & \underline{\textbf{82.7}} & \underline{\textbf{83.7}} & \underline{1092} \\
\hline
\textit{tracking-by-detection} & \multicolumn{7}{c}{} \\
ByteTrack~\citep{bytetrack} & 2022 & 63.1 & 62.0 & 64.5 & 80.3 & 77.3 & 2196 \\
BoT-SORT~\citep{botsort} & 2022 & 65.0 & 65.5 & 64.9 & 80.5 & 80.2 & 1212 \\
OC\_SORT~\citep{ocsort} & 2023 & 63.2 & 63.4 & 63.2 & 78.0 & 77.5 & 1950 \\
ByteTrackV2~\citep{bytev2} & 2023 & 63.6 & 62.7 & 65.0 & 80.6 & 78.9 & 1239 \\
StrongSORT++~\citep{strongsort} & 2023 & 64.4 & 64.4 & 64.6 & 79.6 & 79.5 & 1194 \\
Deep OC\_SORT~\citep{deepocsort} & 2023 & 64.9 & 65.9 & 64.1 & 79.4 & 80.6 & \textit{1023} \\
MotionTrack~\citep{motiontrack_byte_cmc} & 2023 & 65.1 & 65.1 & 65.4 & 81.1 & 80.1 & 1140 \\
SparseTrack~\citep{sparsetrack} & 2023 & 65.1 & 65.1 & 65.3 & 81.0 & 80.1 & 1170 \\
ImprAsso~\citep{imprasso} & 2023 & 66.4 & 66.6 & 66.4 & \underline{\textit{82.2}} & 82.1 & \underline{\textbf{924}} \\
C-BIoU~\citep{cbiou} & 2023 & 64.1 & 63.7 & 64.8 & 81.1 & 79.7 & 1194 \\
ETTrack~\citep{ettrack} & 2024 & 61.9 & 60.5 & & 79.0 & 75.9 & 2188 \\
MoveSORT~\citep{movesort} & 2024 & 62.2 & 60.4 & 64.3 & 79.5 & 75.1 & 2688 \\
DeepMoveSORT~\citep{deepmovesort} & 2024 & 63.2 & 62.8 & 63.8 & 78.7 & 77.3 & 1524 \\
Hybrid-SORT~\citep{hybridsort} & 2024 & 64.0 & & & 79.9 & 78.7 & \\
SMILETrack~\citep{smiletrack} & 2024 & 65.3 & & & 81.1 & 80.5 & 1246 \\
UCMCTrack$^{\triangle}$~\citep{ucmctrack} & 2024 & 65.7 & 66.4 & 65.3 & 80.6 & 81.0 & 1689 \\
BoostTrack+~\citep{boostrack} & 2024 & 66.4 & \textit{67.7} & 65.4 & 80.6 & 81.8 & 1086 \\
BoostTrack++~\citep{boostrackplus} & 2024 & \underline{66.6} & \underline{\textbf{68.0}} & \underline{\textit{65.5}} & 80.7 & \underline{82.2} & 1062 \\

\hline
\textit{end-to-end} & \multicolumn{7}{c}{} \\
MeMOT~\citep{memot} & 2022 & 56.9 & 55.2 & & 72.5 & 69.0 & 2724 \\
MOTR~\citep{motr} & 2022 & 57.8 & 55.7 & 60.3 & 68.6 & 73.4 & 2439 \\
MeMOTR~\citep{memotr} & 2023 & 58.8 & 58.4 & 59.6 & 72.8 & 71.5 & \underline{1902} \\
MOTRv3~\citep{motrv3} & 2023 & 60.2 & 58.7 & 62.1 & 75.9 & 72.4 & 2403 \\
MOTRv2~\citep{motrv2} & 2023 & \underline{62.0} & \underline{60.6} & \underline{63.8} & \underline{78.6} & \underline{75.0} & \\
MOTIP~\citep{motip} & 2024 & 59.2 & 56.9 & 62.0 & 75.5 & 71.2 & \\
\end{tabular}
\caption{Comparison of tracking methods on the MOT17 benchmark. All tracking-by-detection methods use the same private detector --- ByteTrack's YOLOX-X object detection model~\citep{bytetrack}. The best results for each metric are highlighted in bold, and the second-best results for each metric are highlighted in italics. The best results for each paradigm separately are underlined. Methods marked with $\triangle$ employ test set information. The results are sourced from the MOTChallenge~\citep{mot_challenge} public leaderboard, except for Hybrid-SORT, ETTrack, SMILETrack, MOTRv2, MOTRv3, MeMOT, and MoTIP, in which case the results are taken from the paper.}
\label{tab:mot17_benchmark}
\end{table*}

\begin{table*}
\small
\centering
\begin{tabular}{lccccccc}
Method & Year & HOTA & AssA & DetA & MOTA & IDF1 & IDSW \\
\hline
\textit{offline} & \multicolumn{7}{c}{} \\
SUSHI~\citep{sushi} & 2022 & 64.3 & 67.5 & 61.5 & 74.3 & 79.8 & \underline{\textbf{706}} \\
CoNo-Link~\citep{cono_link} & 2024 & \underline{65.9} & \underline{68.0} & \underline{64.0} & \underline{77.5} & \underline{\textit{81.8}} & 956 \\
\hline
\textit{tracking-by-detection} & \multicolumn{7}{c}{} \\
ByteTrack~\citep{bytetrack} & 2022 & 61.3 & 59.6 & 63.4 & 77.8 & 75.2 & 1223 \\
BoT-SORT~\citep{botsort} & 2022 & 63.3 & 62.9 & 64.0 & 77.8 & 77.5 & 1313 \\
ByteTrackV2~\citep{bytev2} & 2023 & 61.4 & 60.1 & 62.9 & 77.3 & 75.6 & 1082 \\
OC\_SORT~\citep{ocsort} & 2023  & 62.4 & 62.5 & 62.4 & 75.7 & 76.3 & 942 \\
StrongSORT++~\citep{strongsort} & 2023 & 62.6 & 64.0 & 61.3 & 73.8 & 77.0 & 1003 \\
MotionTrack~\citep{motiontrack_byte_cmc} & 2023 & 62.8 & 61.8 & 64.0 & \textit{78.0} & 76.5 & 1165 \\
SparseTrack~\citep{sparsetrack} & 2023 & 63.5 & 63.1 & 64.1 & 78.1 & 77.6 & 1120 \\
Deep OC\_SORT~\citep{deepocsort} & 2023 & 63.9 & 65.7 & 62.4 & 75.6 & 79.2 & 779 \\
ImprAsso~\citep{imprasso} & 2023 & \textit{64.6} & 64.6 & 64.8 & \underline{\textbf{78.6}} & 78.8 & 992 \\
MoveSORT~\citep{movesort} & 2024 & 60.5 & 60.0 & 61.3 & 74.3 & 74.0 & 943 \\
DeepMoveSORT~\citep{deepmovesort} & 2024 & 60.6 & 60.8 & 60.5 & 73.6 & 74.1 & 1180 \\
UCMCTrack$^{\triangle}$~\citep{ucmctrack} & 2024 & 62.8 & 63.5 & 62.4 & 75.6 & 77.4 & 1355 \\
C-TWIX~\citep{twix} & 2024 & 63.1 & 62.5 & \textit{64.1} & 78.1 & 76.3 & 5820 \\
SMILETrack~\citep{smiletrack} & 2024 & 63.4 & & & \textit{78.2} & 77.5 & 1208 \\
Hybrid-SORT~\citep{hybridsort} & 2024 & 63.9 & & & 76.7 & 78.4 & \\
BoostTrack+~\citep{boostrack} & 2024 & \textit{66.2} & \textit{68.6} & \textit{64.1} & 77.2 & 81.5 & 827 \\
BoostTrack++~\citep{boostrackplus} & 2024 & \underline{\textbf{66.4}} & \underline{\textbf{68.7}} & \underline{\textbf{64.3}} & 77.0 & \underline{\textbf{82.0}} & \underline{\textit{762}} \\
\hline
\textit{end-to-end} & \multicolumn{7}{c}{} \\
MeMOT~\citep{memot} & 2022 & 54.1 & 55.0 & & 63.7 & 66.1 & \underline{1938} \\
MOTRv2~\citep{motrv2} & 2023 & \underline{61.0} & \underline{59.3} & \underline{63.0} & \underline{73.1} & \underline{76.2} & \\
\end{tabular}
\caption{Comparison of tracking methods on the MOT20 benchmark using a private detector (ByteTrack's YOLOX-X object detection model~\citep{bytetrack}). The best results for each metric are highlighted in bold, and the second-best results for each metric are highlighted in italics. The best results for each paradigm separately are underlined. Methods marked with $\triangle$ employ test set information. The results are sourced from the MOTChallenge~\citep{mot_challenge} public leaderboard, except for Hybrid-SORT, SMILETrack, MeMOT, and MOTRv2, in which case the results are taken from the paper.}
\label{tab:mot20_benchmark}
\end{table*}

\textbf{DanceTrack Benchmark}. A comparison of various tracking methods on the DanceTrack dataset is presented in Table~\ref{tab:dancetrack_benchmark}. Compared to MOTChallenge, we observe a significant shift in the nature of the results. First, considering improvements relative to the strong ByteTrack baseline, the total improvement is $19.8\%$, i.e., from $53.6\%$ to $73.4\%$, in terms of HOTA (MOTRv2), $21.7\%$, i.e., from $55.3\%$ to $76.0\%$, in terms of IDF1 (MOTRv2), and $0.6\%$, i.e., from $92.3\%$ to $92.9\%$, in terms of MOTA (MOTRv3). Since ByteTrack's performance in terms of MOTA is as high as $92.3\%$, the relative improvement is expected to be lower compared to HOTA, which is only $53.6\%$. As shown in the table, these performance gains stem from the introduction of new end-to-end methods. Besides the original MOTR trackers, every end-to-end tracker significantly outperforms all tracking-by-detection methods. Specifically, MeMOTR, which has the lowest performance among end-to-end trackers, still outperforms CoNO-Link, the best-performing tracking-by-detection (online and offline) tracker, by $4.7\%$ in terms of HOTA. Concretely, MOTRv2 achieves the best tracking performance, with $73.4\%$ HOTA, $92.1\%$ MOTA, and $76.0\%$ IDF1. As MOTRv2 is not a \textit{true end-to-end} method, we also highlight the second-best method, MOTIP, with $71.4\%$ HOTA, $91.6\%$ MOTA, and $76.3\%$ IDF1. It is important to note that all tracking-by-detection and offline methods use the same public YOLOX detector~\citep{dancetrack}, while the end-to-end methods train private detectors. This implies that some improvements come directly from the quality of the detections, which can be partially observed through the DetA metric. For an ideal and isolated association performance comparison, it would be necessary to use the DETR detector's outputs instead of the YOLOX detector for tracking-by-detection method evaluation. Unfortunately, this information is not available in the existing literature. 

From Table~\ref{tab:dancetrack_benchmark}, we also observe that SORT-like methods with deep learning-based motion models (e.g., MoveSORT, ETTrack, MotionTrack) prove highly effective on the DanceTrack dataset. Even without appearance features or a rich set of heuristics, these methods improve the DanceTrack benchmark baselines. For instance, compared to ByteTrack, MotionTrack improves performance by $4.6\%$ (i.e., from $53.6\%$ to $58.2\%$) in terms of HOTA. Furthermore, combining improved motion and re-ID models (e.g., Deep OC\_SORT, DeepMoveSORT) leads to even stronger performance. Specifically, compared to ByteTrack, DeepMoveSORT achieves a $9.4\%$ improvement (i.e., from $53.6\%$ to $63.0\%$) in terms of HOTA. The best-performing online tracking-by-detection model is UCMCTrack~\footnote{UCMCTrack requires manually configured camera parameters for each scene, even on the test set.}, which employs video consistent CMC~\citep{ucmctrack} and performs association in a mapped ground plane. The best offline variant is CoNo-Link, a GNN based method with $63.8\%$ HOTA. 

As with MOTChallenge, we discuss the effectiveness of the most important association components of tracking-by-detection methods: motion model, re-ID model, CMC, and association heuristics. In the previous paragraph, we analyzed the effectiveness of learnable deep learning-based motion models~\citep{ettrack, movesort, deepmovesort, motiontrack}. Even for baselines such as SORT and ByteTrack, replacing the KF with stronger motion models improves HOTA by up to $4.8\%$~\citep{deepmovesort} on the DanceTrack validation set. This is expected, as the dancers in this dataset exhibit dynamic, non-linear motion, which the KF fails to model effectively. We also observe the importance of the re-ID model, which strongly influences performance, as demonstrated in the ablations in~\citep{deepocsort, deepmovesort}. In~\citep{deepmovesort} ablation study, it can be seen that the performance of ByteTrack on the DanceTrack validation set improves by up to $5.2\%$ (i.e., from $53.8\%$ to $59.0\%$) in terms of HOTA. At first glance, this is unexpected since the dancers typically wear similar uniforms. While this makes appearance-based association challenging, it is far from obsolete. The use of CMC is noted in~\citep{deepocsort}, but it is not as critical as in the MOT17 benchmark. Strong motion or re-ID models suffice for cases with non-static cameras. Lastly, heuristics are also beneficial on DanceTrack, and their importance is on par with re-ID and motion models~\citep{ocsort, deepocsort, sparsetrack, movesort, deepmovesort, ettrack, hybridsort}.

\begin{table*}
\small
\centering
\begin{tabular}{lcccccc}
Method & Year & HOTA & AssA & DetA & MOTA & IDF1 \\
\hline
\textit{offline} & \multicolumn{6}{c}{} \\
SUSHI$^{*}$~\citep{sushi} & 2022 & 63.3 & 50.1 & 80.1 & 87.7 & 63.3 \\
CoNo-Link$^{*}$~\citep{cono_link} & 2024 & \underline{63.8} & \underline{50.7} & \underline{80.2} & \underline{89.7} & \underline{64.1} \\
\hline
\textit{tracking-by-detection} & \multicolumn{6}{c}{} \\
ByteTrack~\citep{bytetrack} & 2022 & 53.6 & 36.1 & 79.9 & \textit{92.3} & 55.3 \\
BoT-SORT~\citep{bytetrack} & 2022 & 54.2 & 36.8 & 80.0 & 92.1 & 57.0 \\
OC\_SORT~\citep{ocsort} & 2023 & 55.1 & 38.0 & 80.4 & 89.4 & 54.9 \\
SparseTrack~\citep{sparsetrack} & 2023 & 55.5 & 39.1 & 78.9 & 91.3 & 58.3 \\
C-BIoU~\citep{cbiou} & 2023 & 60.6 & 45.4 & 81.3 & 91.6 & 61.6 \\
Deep OC\_SORT~\citep{deepocsort} & 2023 & 61.3 & 45.8 & 81.6 & 91.8 & 61.5 \\
MoveSORT~\citep{movesort} & 2024 & 56.1 & 38.7 & 81.6 & 91.8 & 56.0 \\
ETTrack~\citep{ettrack} & 2024 & 56.4 & 39.1 & 81.7 & 92.2 & 57.5 \\
MotionTrack~\citep{motiontrack} & 2024 & 58.2 & 41.7 & 81.4 & 91.3 & 58.6 \\
C-TWIX~\citep{twix} & 2024 & 62.1 & 47.2 & 81.8 & 91.4 & 63.6 \\
DeepMoveSORT~\citep{deepmovesort} & 2024 & 63.0 & 48.6 & \underline{82.0} & \underline{92.6} & \underline{65.0} \\
UCMCTrack$^{\triangle}$~\citep{ucmctrack} & 2024 & \underline{63.6} & \underline{51.3} & 78.9 & 88.9 & \underline{65.0} \\
\hline
\textit{end-to-end} & \multicolumn{6}{c}{} \\
MOTR~\citep{motr} & 2022 & 54.2 & 40.2 & 73.5 & 79.7 & 51.5 \\
MeMOTR~\citep{memotr} & 2023 & 68.5 & 58.4 & 80.5 & 89.9 & 71.2 \\
MOTRv3~\citep{motrv3} & 2023 & 70.4 & 59.3 & \underline{\textbf{83.8}} & \underline{\textbf{92.9}} & 72.3 \\
MOTRv2~\citep{motrv2} & 2023 & \underline{\textbf{73.4}} & \underline{\textbf{64.4}} & \textit{83.7} & 92.1 & \textit{76.0} \\
MOTIP~\citep{motip} & 2024 & \textit{71.4} & \textit{62.8} & 81.3 & 91.6 & \underline{\textbf{76.3}}

\end{tabular}
\caption{Comparison of tracking methods on the DanceTrack benchmark using the public DanceTrack~\citep{dancetrack} detector. The best results for each metric are highlighted in bold, while the second-best results are highlighted in italics. The best results for each paradigm separately are underlined. Methods marked with $*$ are restricted to offline mode, while those marked with $\triangle$ employ test set information. The results SORT, ByteTrack and BoT-SORT were produced by us. All other results are sourced from their respective published papers.}
\label{tab:dancetrack_benchmark}
\end{table*}

\textbf{SportsMOT Benchmark}. A comparison of various tracking methods on the SportsMOT dataset is presented in Table~\ref{tab:sportsmot_benchmark}. As SportsMOT is a less-known benchmark compared to DanceTrack and MOTChallenge, evaluations for some important methods are missing. Compared to the ByteTrack baseline, we observe a total improvement of $14.6\%$ (DeepMoveSORT), i.e., from $64.1\%$ to $78.7\%$, in terms of HOTA; $2.3\%$ (ETTrack), i.e., from $94.5\%$ to $96.8\%$, in terms of MOTA; and $10.3\%$ (DeepMoveSORT), i.e., from $71.4\%$ to $81.7\%$, in terms of IDF1. In this case, tracking-by-detection methods outperform end-to-end methods by up to $6.8\%$ in terms of HOTA. However, this comparison is not entirely valid, as the YOLOX detector used by tracking-by-detection methods is trained on both the train and validation sets for evaluation on the test set~\citep{sportsmot, deep_eiou, ettrack, movesort, deepmovesort, motiontrack}, whereas end-to-end methods are trained only on the training dataset~\citep{memotr, motip}. 

We discuss the effectiveness of the most important association components of tracking-by-detection methods: motion model, re-ID model, CMC, and association heuristics. In this case, there is a notable difference in performance between ByteTrack, which does not use a re-ID model in this evaluation, and BoT-SORT, which does. The BoT-SORT tracker outperforms ByteTrack by $4.6\%$, i.e., from $64.1\%$ to $68.7\%$. The importance of appearance features is further confirmed through ablation studies in~\citep{deep_eiou, deepmovesort}. Since the motion in SportsMOT is even more dynamic compared to DanceTrack, due to both player and camera movements, tracking methods with deep learning-based motion models prove even more effective. This is evident from the improved performance of MoveSORT, ETTrack, and MotionTrack compared to ByteTrack. Additionally, OC\_SORT, which partially resolves some issues of the KF and thus enhances its capabilities, demonstrates very strong performance despite relying solely on the object detector, the KF, and observation-centric heuristics~\citep{ocsort}. For this domain, CMC methods are not as critical since the camera motion is predictable. Lastly, most association-related heuristics are less effective~\citep{deepmovesort}, primarily due to the sparsity of objects in the scene, where occlusions are less frequent compared to other datasets~\citep{sportsmot}. Notably, bounding box expansion before association proves to be highly effective, as tracks' bounding boxes between consecutive frames often have low IoU due to rapid camera or object motion.

\begin{table*}
\centering
\small
\begin{tabular}{lcccccc}
Method & Year & HOTA & AssA & DetA & MOTA & IDF1 \\
\hline
\textit{tracking-by-detection} & \multicolumn{6}{c}{} \\
ByteTrack~\citep{bytetrack} & 2022 & 64.1 & 52.3 & 78.5 & 95.9 & 71.4 \\
BoT-SORT~\citep{botsort} & 2022 & 68.7 & 55.9 & 84.4 & 94.5 & 70.0 \\
OC\_SORT~\citep{ocsort} & 2023 & 73.7 & 61.5 & \textit{88.5} & 96.5 & 74.0 \\
MixSORT-OC~\citep{sportsmot} & 2023 & 74.1 & 62.0 & \textit{88.5} & 96.5 & 74.4 \\ 
MixSORT-Byte~\citep{sportsmot} & 2023 & 65.7 & 54.8 & 78.8 & 96.2 & 74.1 \\ 
Deep-EIoU~\citep{deep_eiou} & 2023 & \textit{77.2} & \textit{67.7} & 88.2 & 96.3 & \textbf{79.8} \\
MotionTrack~\citep{motiontrack} & 2024 & 74.0 & 61.7 & \underline{\textbf{88.8}} & 96.6 & 74.0 \\
ETTrack~\citep{movesort} & 2024 & 74.3 & 62.1 & \underline{\textbf{88.8}} & \underline{\textbf{96.8}} & 74.5 \\
MoveSORT~\citep{movesort} & 2024 & 74.6 & 63.7 & 87.5 & \textit{96.7} & \textit{76.9} \\
DeepMoveSORT~\citep{deepmovesort} & 2024 & \underline{\textbf{78.7}} & \underline{\textbf{70.3}} & 88.1 & 96.5 & \underline{\textbf{81.7}} \\
\hline
\textit{end-to-end} & \multicolumn{6}{c}{} \\
MeMOTR~\citep{memotr} & 2022 & 70.0 & 59.1 & 83.1 & 91.5 & 71.4 \\
MOTIP~\citep{motip} & 2024 & \underline{71.9} & \underline{62.0} & \underline{83.4} & \underline{92.9} & \underline{75.0} \\
\end{tabular}
\caption{Comparison of tracking methods on the SportsMOT benchmark. The best results for each metric are highlighted in bold, while the second-best results are highlighted in italics. The best results for each paradigm separately are underlined. The results for MoveSORT are taken from~\citep{movesort}, for Deep-EIoU and BoT-SORT are taken from~\citep{deep_eiou}, and all other results are taken from~\citep{sportsmot}}
\label{tab:sportsmot_benchmark}
\end{table*}

\textbf{Multi-Domain Benchmark}. Until recently, the MOTChallenge benchmark, focused on the multi-pedestrian tracking domain, was the primary benchmark for tracker evaluation. It enabled the verification of pivotal tracker methods~\citep{sort, deep_sort, bytetrack}. Moreover, it was the only benchmark for \textit{general} tracking. Trackers are often designed to specialize in a specific domain; for example, trackers~\citep{boostrack, boostrackplus, imprasso, motiontrack_byte_cmc} are evaluated exclusively on MOTChallenge, while the tracker~\citep{deep_eiou} is evaluated only on SportsMOT and SoccerNet~\citep{soccernet}. Results on these benchmarks can sometimes be misleading. For instance, evaluating end-to-end methods only on MOTChallenge could lead to the incorrect conclusion that they generally underperform compared to tracking-by-detection methods. However, this is proven false when comparing performance on DanceTrack.

To evaluate the general tracking performance of different methods, we collect and aggregate their metrics across multiple benchmarks. Since trackers are rarely evaluated on SportsMOT compared to MOTChallenge and DanceTrack, we construct two sets of aggregated results. The first set, presented in Table~\ref{tab:mot17_mot20_dancetrack_macro_benchmark}, includes weighted macro-averaged results with weights $0.5$, $0.5$, and $1.0$ for MOT17, MOT20, and DanceTrack, respectively. Effectively, MOT17 and MOT20 are treated as a single domain. The second set, presented in Table~\ref{tab:mot17_mot20_dancetrack_sportsmot_macro_benchmark}, includes weighted macro-averaged results with weights $0.5$, $0.5$, $1.0$, and $1.0$ for MOT17, MOT20, DanceTrack, and SportsMOT, respectively.

Based on the results presented in Table~\ref{tab:mot17_mot20_dancetrack_macro_benchmark}, we can rank tracking paradigms by performance as follows: end-to-end, offline, and tracking-by-detection. The best-performing tracker is MOTRv2, with $67.4\%$ HOTA and $75.8$ IDF1. Compared to the best offline method, CoNo-Link, MOTRv2 achieves a $2.2\%$ higher HOTA. Compared to the best online tracking-by-detection method, UCMCTrack, MOTRv2 outperforms it by $3.5\%$ in terms of HOTA. Interestingly, ByteTrack performs best in terms of MOTA, achieving a score of $85.7\%$. ByteTrack outperforms MOTRv2 by $1.7\%$ in MOTA. Two other methods with similar MOTA are BoT-SORT and SparseTrack, both of which closely resemble ByteTrack.

Unfortunately, results for \textit{real} end-to-end methods on MOT20 are not available and are therefore missing from this benchmark. Since DETR tends to underperform on crowded images with small objects~\citep{detr, deformable_detr}, we infer poor performance, which is possibly the reason behind the missing results. MOTRv2 stands out, as it mitigates DETR's limitations by using YOLOX for detection proposals.  Lastly, we observe that offline methods based on GNNs outperform online variants within the tracking-by-detection paradigm.

When focusing solely on the online tracking-by-detection results in Table~\ref{tab:mot17_mot20_dancetrack_macro_benchmark}, UCMCTrack emerges as the best-performing tracker in scenarios where camera parameters can be estimated for ground plane projection. The second tier of trackers in terms of performance includes Deep OC\_SORT and DeepMoveSORT, both of which employ a combination of improved motion models and appearance-based association. Other methods fall behind as they lack these critical components.

In Table~\ref{tab:mot17_mot20_dancetrack_sportsmot_macro_benchmark}, we observe only a few tracking-by-detection methods that publish results on all datasets. The best-performing method is DeepMoveSORT, with $67.9\%$ in terms of HOTA and $74.1\%$ in terms of IDF1. ByteTrack is the best tracker in terms of MOTA, achieving a score of $89.1\%$. Unfortunately, many trackers are missing as they lack evaluations on either MOT20 or SportsMOT. However, we present these results as baselines, hoping for more robust evaluations across multiple datasets in future MOT works. We note that a better alternative to this \textit{improvisation} for general tracking evaluation would be a large-scale multi-domain MOT dataset.

\begin{table*}
\centering
\small
\begin{tabular}{lcccccc}
Method & Year & HOTA & AssA & DetA & MOTA & IDF1 \\
\hline
\textit{offline} & \multicolumn{6}{c}{} \\
SUSHI$^{*}$~\citep{sushi} & 2022 & 64.4 & 58.9 & 71.8 & 82.7 & 72.4 \\
CoNo-Link$^{*}$~\citep{cono_link} & 2024 & \underline{\textit{65.2}} & \underline{\textit{59.3}} & \underline{\textit{72.8}} & \underline{84.9} & \underline{73.4} \\
\hline
\textit{tracking-by-detection} & \multicolumn{6}{c}{} \\
ByteTrack~\citep{bytetrack} & 2022 & 57.9 & 48.4 & 71.9 & \underline{\textbf{85.7}} & 65.8 \\
BoT-SORT~\citep{botsort} & 2022 & 59.2 & 50.5 & 72.2 & \textit{85.6} & 67.9 \\
OC\_SORT~\citep{ocsort} & 2023 & 59.0 & 50.5 & 71.6 & 83.1 & 65.9 \\
SparseTrack~\citep{sparsetrack} & 2023 & 59.9 & 51.6 & 71.8 & 85.4 & 68.6 \\
Deep OC\_SORT~\citep{deepocsort} & 2023 & 62.8 & 55.8 & \underline{72.4} & 84.6 & 70.7 \\
MoveSORT~\citep{movesort} & 2024 & 58.7 & 49.4 & 72.2 & 84.4 & 65.3 \\
DeepMoveSORT~\citep{deepmovesort} & 2024 & 62.4 & 55.2 & 72.1 & 84.4 & 70.4 \\
UCMCTrack$^{\triangle}$~\citep{ucmctrack} & 2024 & \underline{63.9} & \underline{58.1} & 71.4 & 83.5 & \underline{72.1} \\
\hline
\textit{end-to-end} & \multicolumn{6}{c}{} \\
MOTRv2~\citep{motrv2} & 2024 & \underline{\textbf{67.4}} & \underline{\textbf{62.2}} & \underline{\textbf{73.6}} & \underline{84.0} & \underline{\textbf{75.8}} \\
\end{tabular}
\caption{Comparison of tracking methods on MOT17, MOT20 and DanceTrack with weights macro-average metric values. The weights for MOT17, MOT20 and DanceTrack are $0.5$, $0.5$ and $1.0$ respectfully. The best results for each metric are highlighted in bold, while the second-best results are highlighted in italics. Methods marked with $\triangle$ employ test set information.}
\label{tab:mot17_mot20_dancetrack_macro_benchmark}
\end{table*}

\begin{table*}
\centering
\small
\begin{tabular}{lcccccc}
Method & Year & HOTA & AssA & DetA & MOTA & IDF1 \\
\hline
\textit{tracking-by-detection} & \multicolumn{6}{c}{} \\
ByteTrack~\citep{bytetrack} & 2022 & 60.0 & 58.5 & 65.4 & \textbf{89.1} & 67.6 \\
BoT-SORT~\citep{botsort} & 2022 & 62.4 & 61.8 & 66.8 & \textit{88.6} & 68.6 \\
OC\_SORT~\citep{ocsort} & 2023 & 63.9 & \textbf{63.2} & 68.2 & 87.6 & 68.6 \\
MoveSORT~\citep{movesort} & 2024 & 64.0 & \textit{62.1} & \textit{69.4} & 88.5 & \textit{69.2} \\
DeepMoveSORT~\citep{deepmovesort} & 2024 & \textbf{67.9} & 60.2 & \textbf{77.4} & 88.4 & \textbf{74.1} \\
\end{tabular}
\caption{Comparison of tracking methods on MOT17, MOT20, DanceTrack, and SportsMOT with weights macro-average metric values. The weights for MOT17, MOT20, DanceTrack and SportsMOT are $0.5$, $0.5$, $1.0$ and $1.0$ respectfully. The best results for each metric are highlighted in bold, while the second-best results are highlighted in italics.}
\label{tab:mot17_mot20_dancetrack_sportsmot_macro_benchmark}
\end{table*}

\section{Conclusion}
\label{sec:conclusion}

This paper presents a comprehensive survey with an in-depth analysis of deep learning-based methods in MOT, covering the field from its foundations to modern state-of-the-art approaches. We review and categorize tracking-by-detection methods into five categories: joint detection and embedding, heuristic-based, motion-based, affinity learning, and offline methods. Additionally, we review end-to-end MOT methods and compare them in detail with tracking-by-detection approaches. All recent methods, starting from the year $2022$, are compared, and their results are analyzed on multiple datasets. Furthermore, the generality of these methods is assessed by evaluating their average performance across different domains.  

We conclude that while heavily engineered heuristic-based methods achieve state-of-the-art performance on highly crowded datasets with linear object motion, such as MOTChallenge, methods that improve association using deep learning models prevail on datasets from other domains, such as DanceTrack and SportsMOT, demonstrating great potential. We primarily refer to methods based on affinity learning (including offline variants using GNNs) from the tracking-by-detection paradigm and all end-to-end methods, as they can learn to associate from data and require minimal hyperparameter tuning during inference.  
Finally, since many papers evaluate their methods on only one or two domains, we are unable to fully assess their generality. Future research would benefit from evaluating methods across multiple datasets to provide a more comprehensive understanding of generalization.  

\section*{Acknowledgements}
I would like to thank Professor Mirjana Maljković Ruzicić from the Faculty of Mathematics, University of Belgrade, for her helpful suggestions and guidance on the writing of this paper.

\appendix

\section{Kalman Filter}
\label{appendix:kalman_filter}

\textbf{Kalman Filter}. The Kalman filter (KF) is a Bayesian filter used to estimate the state of a dynamic system based on three key assumptions: (1) the process and observation models are linear, (2) the process noise—representing uncertainties in the process dynamics—and the measurement (observation) noise are Gaussian with zero mean and known covariance, and (3) the Markov assumption, which states that the future state depends only on the current state and not on past states. These assumptions allow the state estimate to evolve over time while preserving a Gaussian distribution throughout the process. 

In practice, within the domain of multi-object tracking, these assumptions (e.g., known process and measurement models) are often relaxed and supplemented with heuristics and domain knowledge, while still achieving satisfactory results. In the following text, we briefly but formally explain the KF process.

We adopt a similar notation for the Kalman filter equations as in~\citep{kalman_net}. Let $\bm{z}_i$ represent the true state of a tracked object at time $t_i$.  A motion model is used to compute the mean $\hat{\bm{z}}_{i+1}$ and covariance $\hat{\bm{P}}_{i+1}$ of the Gaussian prior as:
\begin{align}
\hat{\bm{z}}_{i+1} &= \bm{F} \tilde{\bm{z}}_i \\
\hat{\bm{P}}_{i+1} &= \bm{F} \tilde{\bm{P}}_i \bm{F}^\top + \bm{Q}
\end{align}
Here, $\bm{F}$ is the state transition matrix, i.e., the motion model, $\bm{Q}$ represents the process noise covariance, and $\tilde{\bm{z}}_i$, $\tilde{\bm{P}}_i$ are posterior estimates from the previous time step. The measurement prior follows a Gaussian distribution with parameters:
\begin{align}
\hat{\bm{x}}_{i+1} &= \bm{H} \hat{\bm{z}}_{i+1} \\
\hat{\bm{\Sigma}}_{i+1} &= \bm{H} \hat{\bm{P}}_{i+1} \bm{H}^\top + \bm{R}
\end{align}
After a new observation is received, the posterior distribution $\mathcal{N}(\tilde{\bm{z}}_{i+1}, \tilde{\bm{P}}_{i+1})$ is refined using:
\begin{align}
\bm{K}_{i+1} &= \hat{\bm{P}}_{i+1} \bm{H}^\top \hat{\bm{\Sigma}}_{i+1}^{-1} \\
\Delta \bm{x}_{i+1} &= \bm{x}_{i+1} - \hat{\bm{x}}_{i+1} \\
\tilde{\bm{z}}_{i+1} &= \hat{\bm{z}}_{i+1} + \bm{K}_{i+1} \Delta \bm{x}_{i+1} \\
\tilde{\bm{P}}_{i+1} &= \hat{\bm{P}}_{i+1} - \bm{K}_{i+1} \hat{\bm{\Sigma}}_{i+1} \bm{K}_{i+1}^\top
\end{align}
where observations $\bm{x}_i$ are assumed to be samples from a normal distribution $\mathcal{N}(\bm{H} \bm{z}_i, \bm{R})$, in which case $\bm{H}$ maps states to observations, and $\bm{R}$ denotes the observation noise covariance matrix. The $\bm{K}_{i+1}$ is the \textit{Kalman gain}, and $\Delta \bm{x}_{i+1}$ denotes the \textit{innovation}. The posterior estimates provide the final state prediction at $t_{i+1}$.

In tracking-by-detection approaches, observations $\bm{x}$ capture noisy measurements of object's position and size, i.e., bounding box outputs from object detection models. The KF prior facilitates data association between detections in consecutive frames, combining predictions with the associated detections via Bayesian inference. During this process, the KF smooths noisy detections, leveraging covariance matrices $\bm{Q}$ and $\bm{R}$, which are usually approximated using domain knowledge and heuristics.

The KF is a staple in tracking-by-detection algorithms~\citep{sort, deep_sort, bytetrack, botsort, ocsort, deepocsort} due to its efficiency in modeling object dynamics and handling noisy observations, delivering satisfactory results on datasets where object motion is predominantly linear~\citep{sort, deep_sort, bytetrack, botsort, ocsort}, such as MOT17 and MOT20 datasets~\citep{mot_challenge}. However, this does not hold in the general case, where the performance of the KF declines compared to deep learning-based motion model alternatives~\citep{dancetrack, motiontrack, movesort, deepmovesort}.

\bibliographystyle{abbrvnat}
\bibliography{main}

\end{document}